%% file: main.tex
\documentclass[sigconf,preprint,authorversion,balance=false]{acmart}

\renewcommand\footnotetextcopyrightpermission[1]{}

\usepackage{tikz}
\usetikzlibrary{positioning}
\definecolor{cbgreen}{RGB}{0,158,115}
\definecolor{cbred}{RGB}{213,94,0}
\usepackage{subcaption}
\usetikzlibrary{trees}
\usepackage{multirow}
\usepackage{natbib}
\usepackage{paralist}
\usepackage[table]{xcolor}
\usepackage{xcolor}
\usepackage{array}
\newcolumntype{M}[1]{>{\centering\arraybackslash}m{#1}}
\usepackage{makecell}
\usepackage{adjustbox}

\usepackage{enumitem}
\usepackage{subcaption}
\usepackage{booktabs}

\usepackage{bbm}

\begin{document}
\fancyhead{}

\title[A Dual Perspective on Synthetic Trajectory Generators]{A Dual Perspective on Synthetic Trajectory Generators: Utility Framework and Privacy Vulnerabilities}

\author{Aya Cherigui}
\orcid{0009-0006-7456-6156}
\affiliation{%
  \institution{Orange Research, Belfort, France}
  \city{Belfort}
  \country{France}}
  \affiliation{%
   \institution{Université Marie et Louis Pasteur, CNRS, institut FEMTO-ST (UMR 6174), F-25000 Besançon, France}
   \city{Besançon}
   \country{France}}
\email{aya.cherigui@orange.com}
\author{Florent Guépin}
\orcid{0009-0008-5098-0963}
\affiliation{%
  \institution{Orange Research, Belfort, France}
  \city{Belfort}
  \country{France}}
\email{florent.guepin@orange.com}
\author{Arnaud Legendre}
\orcid{0009-0001-9512-2355}
\affiliation{%
  \institution{Orange Research, Belfort, France}
  \city{Belfort}
  \country{France}}
\email{arnaud.legendre@orange.com}
\author{Jean-François Couchot}
\orcid{0000-0001-6437-5598}
\affiliation{%
  \institution{Université Marie et Louis Pasteur, CNRS, institut FEMTO-ST (UMR 6174), F-25000 Besançon, France}
  \city{Besançon}
  \country{France}}
\email{jean-francois.couchot@univ-fcomte.fr}

\input{abstract}

\maketitle


\section{Introduction}
\label{sec:intro}
\input{introduction}

\section{Background}
\label{sec:background}
\input{background}

\section{A framework to unify utility evaluation in trajectory generation}
\label{sec:utility}
\input{utility}

\section{Privacy evaluation}
\label{sec:privacy}
\input{privacy}

\section{Related Work}
\label{sec:relatedwork}
\input{related_work}

\section{Conclusion}
\label{sec:conclusion}
\input{conclusion}


\appendix
\input{appendix}
\end{document}

%% file: abstract.tex
\begin{abstract}
Human mobility data are used in numerous applications, ranging from public health to urban planning. Human mobility is inherently sensitive, as it can contain information such as religious beliefs and political affiliations. Historically, it has been proposed to modify the information using techniques such as aggregation, obfuscation, or noise addition, to adequately protect privacy and eliminate concerns. As these methods come at a great cost in utility, new methods leveraging development in generative models, were introduced. The extent to which such methods answer the privacy-utility trade-off remains an open problem. In this paper, we introduced a first step towards solving it, by the introduction and application of a new framework for utility evaluation. Furthermore, we provide evidence that privacy evaluation remains a great challenge to consider and that it should be tackled through adversarial evaluation in accordance with the current EU regulation. We propose a new membership inference attack against a subcategory of generative models, even though this subcategory was deemed private due to its resistance over the trajectory user-linking problem. 
\end{abstract}

%% file: introduction.tex
Human mobility data has become a valuable resource for a variety of applications. Including public health~\cite{kishore2021mobility}, personalized services~\cite{cui2018personalized}, urban planning~\cite{kan2019traffic} and environmental monitoring~\cite{liu2023multi}.

Intrinsically, human mobility data contains sensitive information, including personal routines~\cite{gambs2010show}, religious practices~\cite{franceschi-bicchierai2015redditor} or political affiliations~\cite{painter2021political}. Thus, protection of the private information contained in human mobility datasets has attracted a growing interest~\cite{primault2018long}. It culminates with the introduction of data protection regulations, such as the General Data Protection Regulation (GDPR)~\cite{GDPR} in Europe.

Traditional privacy protection techniques typically operate by transforming the original information, using methods such as aggregation~\cite{abul2008never}, obfuscation~\cite{domingo2012microaggregation} and/or noise addition ~\cite{ghane2019tgm,he2015dpt}. These transformations alter the original information and restrict subsequent data usage, resulting in what can be referred to as \textit{utility degradation}. In general, these two conflicting constraints give rise to a trade-off, commonly known as the\textit{ privacy-utility trade-off}~\cite{jin2022survey,miranda2023sok,buchholz2024sok}. 

The data practitioner often tips the scale in favor of utility within the privacy utility trade-off. This can be explained because in practice, the utility cost of the traditional privacy protection techniques is too high for most of the downstream tasks~\cite{primault2018long,noriega2018mapping}. In response, a new privacy protection technique has emerged in the literature, offering hope for a better privacy-utility trade-off: the generation of synthetic trajectory data~\cite{liu2018trajgans}.

The intuition behind the generation of synthetic trajectory data, is to first learn the original distribution from the sensitive individual information. Second, to draw from the learned distribution synthetic trajectory information that cannot be linked back to any real person. There are a variety of solutions to perform this dual task~\cite{liu2018trajgans, rao2020lstm, chu2024simulating,he2015dpt,cunningham2021privacy}.This paper focuses on advances in deep neural networks as generative models for human mobility data~\cite{sohl2015deep,goodfellow2020generative}. To produce synthetic datasets, deep learning solutions have been shown to produce better convincing results than traditional techniques~\cite{feng2020learning,kapp2023generative}. 

Although there is an extensive literature that proposes a range of architectures to build generative models~\cite{rao2020lstm,feng2020learning,cao2021generating,zhu2023difftraj,chu2024simulating,zhu2024controltraj}, there is no standard way to evaluate their performances. Most of them evaluate their utility through a sample of utility metrics, making comparison between architectures difficult~\cite{rao2020lstm,feng2020learning,zhu2023difftraj}. The distributions of individual mobility datasets are inherently complex, reflecting diverse behaviors and spatiotemporal dependencies~\cite{gonzalez2008understanding,song2010limits,pappalardo2015returners}. Thus, if the pool of the metrics considered is not common, the comparison is almost impossible. 

On the privacy side, in coherence with the findings made by~\citet{buchholz2024sok}, we find that most articles related to trajectory generative models based neural networks do not contain any evaluation of privacy, including at most a discussion~\cite{cao2021generating, zhu2023difftraj}, or even do not mention it outside of the introduction~\cite{chu2024simulating, feng2020learning, zhu2024controltraj}. This lack of privacy evaluation is grounded in the literature in two main arguments. First, the inherent randomness stemming from the sampling of the learned distribution. Second, the inference phase that takes as input pure random noise, that let authors believe in a form of \textit{anonymity by design}. In cases where articles contained privacy evidence~\cite{rao2020lstm}, they based their analysis on the Trajectory User Linking (TUL) problem. However, as introduced by~\citet{najjar2022trajectory}, a TUL solver can be identified as a distance-based metric which has been demonstrated in the recent work done by~\citet{yao2025dcr} to not reflect properly the privacy risk that a model presents. Following their recommendation, membership inference attack should be used to better understand a model's privacy. 

\textit{Contributions.} 
We propose a framework for a systematization of the utility evaluation for synthetic trajectory data in Section~\ref{sec:utility}. Our framework includes
\begin{inparaenum}[(i)]
    \item A multidimensional taxonomy of utility metrics designed to consider all facets of mobility data
    \item A set of guidelines to use our framework, that we put in application to compare three different architectures for a given task.
\end{inparaenum} 
    
We investigate the use of the TUL task as a privacy indicator in Section~\ref{subsec:TUL}. This investigation focuses on a subcategory of generative models that use real data samples at inference time to generate the information. We refer to these as blurring models. We provide evidences, both conceptually and empirically, that solely relying on the TUL privacy indicator can give a false sense of privacy.
    
We introduce a new, to the best of our knowledge, Membership Inference Attack (MIA) against blurring models in Section~\ref{subsec:MIAblurringmodels}.

%% file: background.tex
This section aims to give all the necessary information used throughout the paper. Appendix~\ref{appendix:glossary} (Table~\ref{tab:notations}) gives a general view of the different notation used. 

\subsection{Dataset of human mobility}
\label{subsec:trajectories}
We provide a formal definition for a dataset of human mobility and point out three crucial characteristics. Characteristics that will shed light on the inherent difficulties that arise when working with trajectories. 

\subsubsection{Definition}
The mobility of an individual is typically captured through their spatiotemporal trajectories. A spatiotemporal trajectory is defined as a sequence of chronologically ordered spatiotemporal points, $T=[(p_1,t_1),\dots,(p_i,t_i),\dots,(p_{|T|},t_{|T|})]$, with $|T|$ denoting the length of the trajectory (e.g., the number of visits) and $p$ is the geographic position in a given reference system (e.g., $p=(latitude, longitude)$). Each $t_i$ denotes a timestamp, it is also use to order the sequence. 

Trajectory points can also be enriched with, or reduced to, semantic information, such as visit labels (e.g., shop, restaurant). The temporal component may be omitted, for instance, if observations are regular.

A user's \textit{trace} is the collection of all its trajectories. A dataset composed of traces is termed a \textit{human mobility} dataset. In this paper, we denote such a dataset as $D$, the set of contributing users as $U$, the space of all possible traces as $\Omega$, and the distribution of a dataset of human mobility  as $\mathcal{D}$.

\subsubsection{Key Characteristics}
\label{subsubsec:charact}

Sparsity, heterogeneity and non-stationarity are the three key characteristics we identified. 

\textbf{Sparsity.} Users are not able to move freely in $\Omega$. Road networks, physical constraints or mobility patterns lead to a heterogeneous subset of $\Omega$ and a sparse distribution of trajectories. This leads to mode collapse or poor generalization~\cite{arjovsky2017towards,balestriero2021learning} especially when $\Omega$ is large. 

\textbf{Heterogeneity.} The number of possible trajectories differs significantly from the number of possible paths, indeed there are as many trajectories as different users, for each path. This effect exhibits the diversity of mobility patterns~\cite{pappalardo2013understanding, gonzalez2008understanding, pappalardo2015returners}. The dichotomy proposed by~\citet{pappalardo2015returners} linked distinct mobility patterns to a similar definition of outliers in tabular datasets~\cite{feldman2020neural}.

\textbf{Non-stationarity.} Human mobility presents the paradox to be highly predictable~\cite{gonzalez2008understanding, song2010limits, lu2013approaching} while also being highly context dependent (e.g., traffic congestion or weather conditions). 

\subsection{Data release mechanisms}
\subsubsection{Synthetic trajectory data generation model}
\label{subsubsec:syntheticdata}
Given a human mobility dataset $D$ following an underlying distribution $\mathcal{D}$, a synthetic model is a randomized function $\Phi_D: \Omega \rightarrow \Omega$ trained on $D$, which returns a synthetic dataset $D^s$ from \textit{pure noise} such that $D^s \sim \mathcal{D}$. 
In the rest of the paper, we will use the notation $\mathcal{G}(D) = \Phi_D$ to describe that the procedure $\mathcal{G}$ was trained on $D$.

\subsubsection{Blurring model}
\label{subsubsec:blurringmodels}
In contrast to a synthetic model, we here introduce a new definition for a subset of the synthetic model: a \textit{blurring model} is a randomized function $\tilde{\Phi}_D: \Omega \rightarrow \Omega$ trained on $D$. It does not generate data from noise; instead, it processes a set of real samples $Q$ (drawn from $\mathcal{D}$) to produce a blurred dataset $Q^s$. 
The objective is for the blurred data to retain the original distribution (such that $Q^s \sim \mathcal{D}$). In the rest of the paper we denote the blurring procedure as $\tilde{\mathcal{G}}$, where $\tilde{\mathcal{G}}(D) = \tilde{\Phi}_D$ to describe that the blurring procedure $\tilde{\mathcal{G}}$ was trained on $D$.

When studying the privacy implications of blurring models, two distinct levels must be considered.
The first, focus of most of the privacy literature, concerns the \textit{training dataset $D$} and whether the model reveals information about the data used to train it. 
The second level, specific to blurring models, concerns the \textit{inference dataset $Q$} and whether information from the real records used as input is leak into the generated output. 
Unlike synthetic models, which generate from noise, blurring models inherently carry this second risk because they directly transform real records, maintaining by design a 1:1 user correspondence at inference. 
This distinction makes inference-time privacy crucial, as the confidential input itself is the foundation of the generation process.

\subsubsection{Synthetic data generation with neural networks}
\label{subsubsec:neuralmodels}
Some classes of deep neural networks specialize in learning the underlying distribution of their training set to subsequently generate new data from pure noise, they are called generative models. We now present the deep generative models that serve as a basis for this paper. 

\textbf{Generative Adversarial Networks (GANs).} 
Introduced by \citet{goodfellow2014generative} and revised later~\cite{goodfellow2020generative}, GANs use an adversarial training process between two networks. The first, \textit{the generator}, synthesizes records. In the original version of~\citet{goodfellow2014generative}, it synthesized records from a vector of pure noise. The second, \textit{the discriminator}, has to distinguish the generator's output from real samples. At training time, the discriminator's loss is backpropagated to the generator, which aims to "fool" the discriminator and, in doing so, learning the underlying distribution $\mathcal{D}$. Originally used for image generation~\cite{radford2015unsupervised, arjovsky2017wasserstein}, GANs are now applied to text~\cite{gulrajani2017improved}, tabular data~\cite{xu2019modeling}, and trajectory generation~\cite{rao2020lstm}. About the former,~\citet{liu2018trajgans} first identified and discussed the potential of GANs to generate synthetic trajectory data, in a framework presentation called TrajGAN. They highlight two key remarks associated with using GANs for trajectory generation: 
\begin{inparaenum}[(i)]
    \item The necessity for the model to encoding the input trajectories to create dense representations, for which they suggest Recurrent Neural Networks (RNNs)
    \item Warn against the risk of overfitting, which simultaneously impairs the model's ability to generalize beyond training examples and compromises the anonymity of the original records.
\end{inparaenum}

\begin{table*}[htbp!]
\centering
\begin{tabular}{ccccc}
\toprule
     \multirow{2}{*}{Level}
    & \multicolumn{2}{c}{\textbf{Statistics preservation}} 
    & \multirow{2}{*} {\textbf{Realism assurance} }
    & \multirow{2}{*} {\textbf{Task performance}} \\
     & \textit{Marginal statistics} & \textit{Relational statistics} & & \\
\midrule
 \multirow{4}{*}{\rotatebox{90}{Trajectory}} & Average speed~\cite{zheng2015trajectory} & Hausdorff distance~\cite{rockafellar317jb} & Reachability~\cite{miller2005measurement} & Trajectory clustering~\cite{bian2018survey} \\
& OD spatial density~\cite{chu2024simulating, feng2020learning} & Dynamic time warping \cite{berndt1994using} & Map reconstruction~\cite{wang2020protecting} & Trajectory forecasting~\cite{gambs2012next, rudenko2020human} \\
& I-rank~\cite{gonzalez2008understanding,pappalardo2013understanding} & Fréchet distance \cite{alt1995computing}  & Time reversal ratio \cite{merhi2024synthetic}   \\
& Waiting time~\cite{brockmann2006scaling,gonzalez2008understanding} &&\\
\cmidrule(lr){1-5}
 \multirow{3}{*}{\rotatebox{90}{Point}} & G-rank~\cite{ouyang2018non} & Transition probabilities~\cite{gambs2010show} & Location implausibility~(\ref{appendix:location_implaus}) & Traffic flow prediction~\cite{du2019deep}\\
& Temporal activity~\cite{rao2020lstm} & Spatial co-occurence~\cite{cressie2015statistics} & Semantic implausibility~(\ref{appendix:semantic_implaus}) & Crowd density prediction~\cite{li2019densely}\\
& Categorical G-rank~\cite{bindschaedler2016synthesizing, ouyang2018non, rao2020lstm}& \\
\bottomrule
\end{tabular}
\caption{Overview of utility metrics classified according to the proposed taxonomy.}
\label{tab:taxonomy}
\end{table*}

\textbf{Diffusion Probabilistic Models (DPMs).} 
Introduced by~\citet{sohl2015deep}, DPMs leverage principles from non-equilibrium thermodynamics in a two-part framework. The first, the \textit{forward diffusion process}, transforms one real distribution into another noisy one. The second, the \textit{backward diffusion process}, reverses this transformation to synthesize data. This backward process, analogous to a GAN's \textit{generator}, is used to generate synthetic data from pure noise.~\citet{sohl2015deep} originally proposed DPMs as a general-purpose framework capable of learning any underlying data distribution.

%% file: utility.tex
In this section, we first provide insights about the synthetic trajectory data evaluation methods from the literature, highlighting the need for a systematic approach. In a second part, we provide the basis to answer this challenge by providing a taxonomy of utility metrics. In a third part, we will introduce a unified evaluation framework based on this taxonomy, based on a careful selection of metrics from the taxonomy and a scoring vector to allow a fair comparison between models and use-cases. Finally, we illustrated the use of our framework through two different use-cases. 

\subsection{On the necessity to evaluate the utility of synthetic trajectory data}
Evaluation of the quality of modeled spatio-temporal trajectory data is a complex task. We want to emphasize how the evaluation of trajectory data is a multi-aspect task, where each metric evaluates one side -but not necessarily all- of the utility. In practice, practitioners and researchers often employ a non-standardized sets of metrics, tailored to answer specific applications without providing a comprehensive evaluation of the generated traces. Prior work~\cite{miranda2023sok} introduced a distinction between data preservation (e.g., how close the output data are to the original one), statistics preservation (e.g., the preservation of specific properties of the dataset) and realism assurance (e.g., how realist the generated traces are). While this is a first attempt to a comprehensive taxonomy of utility, we think it misses the applicability of generated traces, by not classifying the ability to infer information through the training of ML models. \citet{buchholz2024sok} introduced the distinction between point and trajectory level while building on the data and statistics preservation introduced by\citet{miranda2023sok}. We think the distinction introduced by~\citet{buchholz2024sok} could be extended for realism assurance, and for what we will call task performance later. 

We introduce a systematic, multi-aspect taxonomy to enable a comprehensive framework dedicated to facilitate the effective adoption of synthetic trajectory data in both industry and research. We believe that fair benchmarking is conditioned by the existence of a consensus about evaluation of utility because without it, we may obscure the model selection process to practitioners. 

\subsection{Taxonomy for utility metrics}
Following the intuitions introduced by~\citet{buchholz2024sok} and~\citet{miranda2023sok}, our taxonomy is structured along two orthogonal axes, as shown in Table~\ref{tab:taxonomy}: \\
(1) The first axis distinguishes metrics based on the data unit they characterize: either at the \textit{Trajectory level}, where entire trajectories are assessed as single units, or at the \textit{Point level}, where individual points, pooled across all trajectories, are evaluated, without regard to the trajectory context. \\
(2) The second axis distinguishes among three utility notions that a synthetic mechanism should aim to maximize. The first two followed the work conducted by~\citet{miranda2023sok}, the \textit{statistics preservation} and the \textit{realism assurance}. Upon these two notions, we introduced a last axis of evaluation, namely the \textit{task performance}, in order to encompass performances variations when performing downstream tasks (e.g. training machine learning models) on original or synthetic data. Notably, we chose to omit the notion of \textit{data preservation}, introduced by~\citet{miranda2023sok} from the taxonomy: It appeared to fit mostly in the category of relational statistics, taken at trajectory level; but moreover, one major conclusion of our study is that relational statistics performed across training and synthetic datasets may lead to misleading interpretations and, most importantly, to a misread level of anonymity.

We will now provide more details on our taxonomy and answer how we can classify each utility metric into each utility notion, for both trajectory and point levels.

\subsubsection{Statistics Preservation}
A statistic is formally defined as a function of the observed data, used to summarize or describe the dataset.
We distinguish between two types of statistics: \textit{Marginal statistics} and \textit{relational statistics}.

\textit{Marginal statistics} describe all metrics that relate to individual trajectory units, taken independently of the other individual trajectory units. At the trajectory level, it includes average speed~\cite{zheng2015trajectory}, origin/ destination spatial density \cite{chu2024simulating, feng2020learning}, I-rank (i.e., the rank of visitation frequencies for an individual's mobility)~\cite{gonzalez2008understanding, pappalardo2013understanding} or waiting time distribution~\cite{brockmann2006scaling,gonzalez2008understanding} (i.e., the time spent in a location by an individual before moving to the next one). At the point level, it includes G-rank (i.e., an indicator of the popularity of the locations)~\cite{ouyang2018non}, temporal activity distribution (e.g., peak hours at a location)~\cite{rao2020lstm} or semantic importance (e.g., the total stay duration in a location)~\cite{bindschaedler2016synthesizing,ouyang2018non,rao2020lstm}.

\textit{Relational statistics} by contrast, describe all metrics that relate to any combination of trajectory units. At the trajectory level, it includes measures of similarity (or alignment) between trajectories such as dynamic time warping~\cite{yi1998efficient} and Hausdorff distance~\cite{atev2010clustering,rockafellar317jb}. These metrics are computed between all possible pairs of trajectories within the dataset, to quantify overall similarity patterns. For a comprehensive list of trajectory similarity metrics, we refer the reader to the work by~\citet{su2020survey}. It also includes any metric related to dataset properties as it refers to all trajectories of a dataset taken together.
At the point level, it includes transition probabilities~\cite{gambs2010show} (i.e., the likelihood of transition from a point in the spatial domain to other points) or spatial co-occurrence between points~\cite{cressie2015statistics} (i.e., the frequency with which two or more spatial points are found in proximity).

\subsubsection{Realism Assurance}

First seen in the work of~\citet{miranda2023sok}, this notion covers all metrics that relate to the evaluation of traces realism (i.e., respect of real-world constraints). It is important to note that generated trajectories that preserve statistics do not imply their realism; As such, it must be independently assessed (e.g., being able to verify the correlation within trajectories is not related to the evaluation of not being able to drive in forest). This distinction follows the intuition used in the work done by~\citet{guan2024zero}, where the generation of aggregated trajectories (macro-information) is unrelated to the realism of the trajectories (micro-information). In their work, they used a probabilistic mobility model~\cite{farzanehfar2021risk} to generate traces that do not aim to be convincing and realistic, but when aggregated together, it produces accurate statistics.

At the trajectory level, it includes reachability~\cite{miller2005measurement}(i.e., evaluation of the feasibility of the entire synthetic trajectory, given the real spatial and temporal constraints), map reconstruction~\cite{wang2020protecting, biagioni2012inferring} (i.e., the reconstruction of road networks from trajectory data) or the time reversal ratio~\cite{merhi2024synthetic}(i.e., how coherent is the temporal consistency, such as a chronological order of timestamps). At the point level, it includes location plausibility~(\ref{appendix:location_implaus}) (i.e., we assess the plausibility of the presence of specific points in the trajectory) or semantic plausibility~(\ref{appendix:semantic_implaus}). 

\subsubsection{Task Performance}
This utility notion considers the effectiveness of synthetic data for downstream applications. Note that data can be collected to serve a specific set of goals, that can be defined apriori by the practitioner. It is hence relevant to address the ability of the synthetic dataset to serve these goals, with a set of metrics dedicated to evaluating the performance of these downstream functions or tasks. These metric results can eventually be compared to those of the original data.
At the trajectory-level, it includes metrics such as trajectory clustering~\cite{bian2018survey}, and trajectory forecasting~\cite{rudenko2020human}. At the point level, it includes traffic flow~\cite{du2019deep} (i.e., the prediction of traffic over the road network based on the seen mobility) and crowd density prediction~\cite{li2019densely}. We refer the reader to the work achieved by~\citet{luca2021survey} for a comprehensive list of human mobility tasks).

\subsubsection{Methodological note: relational statistics and cross-dataset comparisons}
\label{subsubsec:note}
When evaluating relational statistics, a critical methodological choice arises: should we compare distributions computed within each dataset (intra-dataset), or compute metrics across datasets (inter-dataset)?
In the intra-dataset paradigm, we compute the metric independently on the synthetic dataset and on the real dataset, then compare the resulting distributions. For instance, when evaluating trajectory similarity via the Hausdorff distance, we would compute the distribution of pairwise distances within the synthetic dataset, and separately within the real dataset, then compare these two distributions using a distance measure (e.g., Wasserstein distance).
In contrast, the inter-dataset paradigm computes metrics directly between real and synthetic records —for example, measuring the distance from each synthetic trajectory to its closest real counterpart. This approach is exemplified by the \textit{data preservation} category introduced by ~\citet{miranda2023sok}.

We strongly recommend against using the inter-dataset paradigm for utility evaluation, as it conflates utility assessment with privacy measurement. Indeed, metrics that quantify proximity between real and synthetic records are precisely those used to detect memorization and assess privacy risks (e.g., membership inference attacks based on the distance to closest record). Relying on inter-dataset metrics for utility evaluation may create a confusion between the evaluation of quality of the generated traces and their evaluation of privacy. As we demonstrate in Section~\ref{sec:privacy}, such inter-dataset distances can be exploited directly by an attacker to perform membership inference attacks and expose the privacy vulnerabilities of models. In particular, for blurring models that transform real records at inference time. That is why, throughout this paper, we adopt the intra-dataset paradigm: all relational statistics are computed independently within each dataset, and only the resulting distributions are compared.

\subsection{Framework of utility evaluation}
Building on our proposed taxonomy for utility metrics, our goal is now to introduce a framework that will encompass all aspects of utility into one \textit{utility-vector}. This vector will later be used to compare models on specific use-cases. As such, we will consider a utility-vector with one dimension assigned to each member of our resulting matrix, leading to a 8-dimensional vector. 

First, we will present the general mechanism of our framework, and how we can select each metric along each different criterion included in our taxonomy. To provide insights on the application of our framework, in a second part we will provide two different real-world application examples, and show how our framework helps to make an informed decision. 

\begin{table*}[!htbp]
\centering
\begin{tabular}{
    l
    l
    c
    c
    c
    c
}
\toprule
    & \textit{Weekly FS-NYC}
    & \multicolumn{2}{c}{\textbf{Statistics preservation}}
    & \multirow{2}{*}{\textbf{Realism assurance} }
    & \multirow{2}{*}{\textbf{Task performance}} \\
    
    & \textit{(Use-case A)}
    & \textit{Marginal statistics}
    & \textit{Relational statistics}
    & & \\
    \midrule
    \multirow{6.5}{*}{\rotatebox{90}{Trajectory level}}
    &
    & I-rank
    & Cosine similarity
    & Trajectory implausibility
    & Trajectory clustering \\
    & & ($W_1$) & ($W_1$) & (Ratio) & (Silhouette score) \\
    & \textbf{Original}  & 0.000  & 0.000  & 0.075 & 0.595 \\
    \cmidrule(lr){2-6}
    & \textbf{LSTM-TrajGAN~\cite{rao2020lstm}}   & 0.078  &  0.006  & 0.563 & 0.174 \\
    & \textbf{exGAN~\cite{song2023except}}     & 0.120    & \textbf{ <0.001}  &0.613 & \textbf{0.543} \\
    & \textbf{TrajGDM~\cite{chu2024simulating}}     & \textbf{0.015}   & N/A  & \textbf{0.321} & 0.393 \\
\midrule
    \multirow{6.5}{*}{\rotatebox{90}{Point level}}
    &
    & Categorical G-rank
    & Transition probabilities
    & Category-location match
    & Global flow prediction \\

    &  & ($\tau_b$)   & ($W_1$) & (Ratio) & (Mean $W_1$) \\
    & \textbf{Original}  & 1.000 & 0.000 & 1.000 & 0.015 \\
    \cmidrule(lr){2-6}
    & \textbf{LSTM-TrajGAN~\cite{rao2020lstm}}  & 0.511 & 0.190   & 0.257   & \textbf{0.026}   \\
    & \textbf{exGAN~\cite{song2023except}}     &  \textbf{0.822} & \textbf{0.171} &  \textbf{0.260}  & 0.029\\
    & \textbf{TrajGDM~\cite{chu2024simulating}}   & N/A  & 0.611 &  N/A   & 0.052  \\
\bottomrule
\end{tabular}
\caption{Results of the application of our framework on the generated data (Weekly FS-NYC)~\cite{may2020marc}.}

\label{tab:usecaseA}
\end{table*}

\subsubsection{Framework and metric selection guidelines}

For each criterion contained within our taxonomy, we advise the practitioner wanting to compare generative models, to select at least one metric. Then, the practitioner can compute the benchmark of the selected models to compare them. First, models need to be trained with the private human mobility dataset, and the benchmark is then done over the generated trajectory datasets. Each metric provide a score, and all the scores create what we call a \textit{utility-vector}. Comparison of the resulting utility-vectors, such as tailoring the comparison to their use-cases, is left to the practitioner. 

Following the structured overview provided by our taxonomy, the selection process of each metric within a given criterion requires context-aware decision-making. Indeed, the relevance of each metric may vary depending on operational objectives. We identified three different axes that help select each metric within each criterion: the intended application of the synthetic data, the structure of the dataset, and the required level of physical realism. 

The relevance of each metric depends mainly on the primary use-case of the evaluation. Even within categories all metrics might not give the same kind of information about the category under evaluation, this is illustrated in Table~\ref{tab:taxonomy}. As an example, Hausdorff and Frechet metrics are in the same category, and while the former informs on the distance between sets of points, the latter informs on the distance between the two curves of the trajectories.

The structure and granularity of the human mobility dataset may constrain the applicability of certain metrics. It is fairly evident, for instance, that metrics requiring continuous spatial or temporal information (e.g., speed and reachability) are unsuitable for datasets lacking such attributes, whereas richer datasets - such as those including categorical information about travel type - may support a broader range of metrics (e.g., travel type prediction). 

The necessity for physical plausibility varies across applications. In regulatory or safety-critical contexts, realism assurance metrics (e.g., reachability and adherence to road networks) are essential, whereas in analytical or simulation-based studies, minor deviations from realism may be acceptable if key statistical properties are maintained. In practice, this means that one may consider the time to be of lesser importance than the adherence to the road networks, leading to a selection of the latter metric for this criterion. 
To illustrate how one can apply our framework in practice, we will provide now two real life example.

\begin{table*}[!htbp]
\centering
\begin{tabular}{
    l
    l
    c
    c
    c
    c
}
\toprule
    & \textit{Geolife daily}
    & \multicolumn{2}{c}{\textbf{Statistics preservation}}
    & \multirow{2}{*}{\textbf{Realism assurance} }
    & \multirow{2}{*}{\textbf{Task performance}} \\
    
    & \textit{(Use-case B)}
    & \textit{Marginal statistics}
    & \textit{Relational statistics}
    & & \\
    \midrule
    \multirow{6.5}{*}{\rotatebox{90}{Trajectory level}}
    &
    & Average speed
    & Pairwise Hausdorff
    & Map reconstruction
    & Next location prediction \\
    & & ($W_1$ (km/h)) & ($W_1$) & (Mean (km)) & (Mean acc) \\
    & \textbf{Original}  & 0.000  & 0.000 & 0.051 & 0.513  \\
    \cmidrule(lr){2-6}
    & \textbf{LSTM-TrajGAN~\cite{rao2020lstm}}   & 1.163  &  \textbf{0.002} &  \textbf{0.128}  & 0.130 \\
    & \textbf{exGAN~\cite{song2023except}}     & \textbf{0.689} & \textbf{0.002}  & 0.133 & \textbf{0.143} \\
    & \textbf{TrajGDM~\cite{chu2024simulating}}  & N/A   & 0.175  & 1.673 & 0.050 \\
\midrule
    \multirow{6.5}{*}{\rotatebox{90}{Point level}}
    &
    & G-rank
    & Transition probabilities
    & Location implausibility
    & Global flow prediction \\

    &  & ($\tau_b$) & ($W_1$) & (Ratio) & (Mean $W_1$) \\
    & \textbf{Original}  & 1.000 & 0.000 & 0.003 & 0.008 \\
    \cmidrule(lr){2-6}
    & \textbf{LSTM-TrajGAN~\cite{rao2020lstm}}  & \textbf{-0.001} & \textbf{0.157}   & \textbf{0.020}   & \textbf{0.012} \\
    & \textbf{exGAN~\cite{song2023except}}     &  \textbf{-0.001} & 0.172 & \textbf{0.020}  & \textbf{0.012} \\
    & \textbf{TrajGDM~\cite{chu2024simulating}}   & -0.289 & 0.652 &  0.124 & 0.032 \\
\bottomrule
\end{tabular}
\caption{Results of the application of our framework on the generated data (Daily Geolife).}

\label{tab:usecaseB}
\end{table*}
\subsubsection{Application of the framework: two real life examples}

In the rest of the paper, we will refer to the following use-cases by their given label in the following enumeration. \\
\textbf{A.} In this use-case, the data holder is a bank. The dataset is made of the description of the time and location where each user used its credit card, compiled by user over a week. Thus, each point is a combination of: 
\begin{enumerate}[label=\arabic*.]
    \item \textbf{The timestamp} of the purchase.
    \item \textbf{The localization} (longitude and latitude) of the purchase
    \item \textbf{The amount} of money spent for the purchase. 
    \item \textbf{The category} of the store the user spend money in. 
\end{enumerate}
Given this information, the bank wants to train a model to predict market fluctuation locally. Before training their model, they need to ensure their data will be protected, so they want to evaluate which generative model is the best to generate the information it need. They will give priority to the correctness of the categorical information (type of store and amount spend) over the spatial information. \\
\textbf{B.} In this use-case, the data holder is an internet provider. The dataset is made of the description of internet research made by users, compiled by user over a day. Thus each point is a combination of: 
\begin{enumerate}[label=\arabic*.]
    \item \textbf{The timestamp} of the search.
    \item \textbf{The localization} (longitude and latitude) of the search.
    \item \textbf{The category} of the research made by the user. 
\end{enumerate}
Given this information, the internet provider wants to deploy a targeted advertisement model which will send ads related to the predicted location of the user at a given time. Thus, they will give priority to a synthetic model that will be able to provide great performances over every metric linked to the location of the user within the day. 

For both use cases, we will now present our framework application, and how we can draw conclusions on the model to be used.
\paragraph{Metric selection} It depends both on the dataset's characteristics and the goal of the data owners. In scenario (A), the data owner aims to maximize the plausibility of the category, while in scenario (B) the data owner aims to maximize the quality of the overall locations. Keeping this in mind, we here suggest two sets of 8 metrics, alongside their justifications.
\begin{enumerate}[label=\Alph*.]
    \item At the trajectory level:
    \{I-rank, Cosine Similarity, Trajectory implausibility, Trajectory clustering\}. At the point level: \{Category-location match, Transition probabilities, Category-location match, Crown density prediction\}. The motivation behind this choice is to maximize the probability of generating correct categorical information, according to the data owner's priorities. 
    \item At the trajectory level: \{Average speed, Hausdorff distance, map reconstruction, trajectory forecasting\} and at the point level: \{G-rank, transition probabilities, location implausibility, traffic flow prediction\}. The motivation behind this choice is to maximize the probability of generating correct trajectory information, according to the data owner's priorities.
\end{enumerate}
For further development over the specificity of each metric, we refer the reader to Appendix~\ref{appendix:utilitymetrics}.

\paragraph{Experimental setup.} The two chosen use-cases can be illustrated through the two publicly available datasets presented in the next paragraph. Weekly Foursquare~\cite{may2020marc} is made of a succession of check-ins, that can be seen as a succession of purchases, that contains the necessary information about our use-case (A). Similarly, our processed version of Geolife~\cite{zheng2008understanding, zheng2009mining, zheng2010geolife}, centered on Beijing and keeping traces over a range time of one day, contains the information to fulfill the need of our use-case (B). In both cases, we applied a split 2/3, 1/3 on the datasets to build the training and validation/testing sets. For both use-cases, we trained three different generative models, as presented in~\ref{subsubsec:neuralmodels} and aim to select the more suitable one for each use-case. We trained the models using a GPU Tesla T4 with 16GB of VRAM and a GPU NVIDIA L40S with 48GB of VRAM. More details about the models and their hyperparameters can be found in Appendix~\ref{appendix:training_setup}. All subsequent computations were done with a CPU dual Interl(R) Xean(R) Gold 6438Y+ with 64 hearts.

\paragraph{Models and datasets}
To favorize diversity in our experiment, we choose three different models from the literature, two GANs and one DPM, that we trained over two different publicly available datasets, specifically chosen to illustrate our two use-cases. Models we will use in this paper are presented below.

\textbf{LSTM-TrajGAN~\cite{rao2020lstm}.} LSTM-TrajGAN builds on the TrajGAN framework~\cite{liu2018trajgans} to jointly model spatial, temporal, and semantic attributes. In this architecture, the generator begins by embedding each of the features using Multilayer Perceptrons (MLPs), then concatenates them with a noise vector to introduce randomness, and last fuses them into a fixed-sized latent representation using a Fully Connected (FC) layer. This resulting latent representation is then passed through a many-to-many Long-Short Term Memory (LSTM) layer~\cite{hochreiter1997long}, capturing the sequential dependencies across trajectory points and generating outputs with the same timestamps as the input. Second, the discriminator processes the generated trajectories similarly to the generator during the embedding and fusion phases but employs a many-to-one LSTM to aggregate them into a prediction. In this architecture, the loss is a function called TrajLoss that includes all dimensionalities (e.g., spatial, temporal and categorical) of the trajectories. 

\textbf{exGAN~\cite{song2023except}.} Many LSTM TrajGAN variants have been proposed in the literature~\cite{song2023except, shin2023tcac}. One of them is an architecture called exGAN. exGAN architecture is mostly similar to the LSTM-TrajGAN architecture, as it follows the exact same generation procedure, but it differs in the discriminator. To introduce an except-condition to exclude sensitive labels. The architecture also replaces the LSTM module with attention layers to enhance performance and prevent label leakage.

\textbf{TrajGDM~\cite{chu2024simulating}.} This diffusion architecture preprocesses each trajectory as a sequence of discrete spatial units. To do so, each trajectory goes into an LSTM-based encoder, that encodes it into a continuous latent space. The location encoding function incorporates spatial adjacency relationships between locations. In the forward process, noise is added incrementally to the encoded trajectories. In the reverse process, a transformer-based network iteratively estimates and removes the added noise to reconstruct the trajectories. The network combines a LSTM for sequential continuity and transformer layers for long-range spatial-temporal dependencies. Finally, the refined representation is decoded into discrete trajectory points using a transformer-based decoder, which outputs the probability distribution of candidate locations in an auto-regressive manner (eg. generation of sequences by timestamps $t$).

For metrics that require spatial discretization such as density-based metrics, G-rank or PoI-category match, grid size selection is done by the same pre-specified rule in Appendix~\ref{appendix:Discretization}. More details on the metric calculation are in appendix~\ref{appendix:utilitymetrics}

\label{subsubsec:datasets}
To illustrate the two use-cases previously introduced, we have chosen two publicly available datasets. Our choice was led by what we think a proprietary dataset for those use-cases would look like. We trained our three different models using those two datasets.

\textbf{Weekly Foursquare.} Originally used in the LSTM-TrajGAN paper~\cite{rao2020lstm}, Weekly Foursquare (Weekly FS-NYC) is an extraction from the Foursquare NYC check-ins dataset~\cite{yang2014modeling}, an extraction provided by~\citet{may2020marc}. They kept the user ID, trajectory ID, location, hour, day and category attributes over all the original attributes. The extracted datast is constituted of 3 079 trajectories over 193 users.

\textbf{Geolife.} Introduced and maintained by~\citet{zheng2008understanding,zheng2009mining,zheng2010geolife}, the original Geolife dataset is constituted of 17 621 trajectories collected between April 2007 and April 2012, across 182 users. Those trajectories are made of GPS points, with a high sampling rate, leading to dense trajectories. We preprocessed Geolife to vary the sampling rate between 3 different format, Weekly, Daily and Hourly. Details about the Geolife preprocessing can be found in Appendix~\ref{appendix:properties}. Due to space constraints, we decided to keep only the Daily version -which we believed fit the best our use-case B- of the preprocessed Geolife in the main paper. Tables~\ref{tab:usecaseBWG} and~\ref{tab:usecaseBHG} in Appendix~\ref{appendix:add_datasets} present results computed over the Weekly and the Hourly versions of Geolife.

\textbf{Results and interpretation.} We now aim to interpret the results provided by our framework of utility we are doing so by putting ourselves in the practitioner/ data owner perspective for both use-cases. In both use-cases, we have chosen metrics in the framework that we believed assess best the features we wanted to maximize utility on. 

In use-case A, we can observe that each model performed better for some specific tasks. Table~\ref{tab:usecaseA} presents a compilation of the three utility-vectors (e.g., the utility-vector for LSTM-TrajGAN can be extracted from the table as (0.078, 0.006, 0.563, 0.174, 0.511, 0.190, 0.257, 0.026)). For I-rank and trajectory implausibility TrajGDM performed better, for global flow prediction LSTM-TrajGAN outperformed the other models, and for all the other metrics, exGAN achieved greater success than the other models. We here naively consider the number of categories best computed by each model, leading to the choice of exGAN. We can notice that taken in consideration the variety of application on practitioner may face, their perspective may lead to a different choice. The diversity of the results in that use-case could also lead the practitioner to train a fourth, categorical specific, model, to outperform all three other models. For example, the practitioner could choose to mix TrajGDM and exGAN, to generate, respectively, the trajectory and the categorical information. 

In use-case B, Table~\ref{tab:usecaseB} presents the compilation of the three utility-vectors. From them, it can be seen that LSTM-TrajGAN and exGAN outperformed significantly TrajGDM, on every single metric selected in the framework. Although those results are highly dependent on the dataset at hand and the models selected, they are very significant for the practitioner. Both LSTM-TrajGAN and exGAN are very close to the original dataset in term of utility. They often achieve the same performances even, as for the Pairwise Hausdorff metric, the location implausibility, the G-rank and the global flow prediction. This finds an explanation in the closeness of their architectures. Overall, while both models are providing accurate results, map reconstruction and transition probabilities could have been seen as more important by our considered practitioner for use-case B, as those metric are mapping the realisticness of the synthetic world created. Overall, if the practitioner would ground its choice \textit{only} over the utility consideration, LSTM-TrajGAN would be the model to select. 
This can be explained visually (as per the goal to generate accurate trajectory information) because both models are providing similar, translated and blurred version of the original trajectories. 

In both use-cases A and B, the practitioner, respectively, would have selected exGAN and LSTM-TrajGAN -two blurring models- if the selection was based only on the utility. In the rest of the paper, we will demonstrate that this choice is compromised (by anonymity assessment) by the propensity of blurring models to memorize their training dataset.

We will make the code available on Github after publication for reproducibility, extensibility and application of the framework. 

%% file: privacy.tex
In this section, we discuss the privacy evaluations made in the different articles found in the literature. We motivate this part by the dual role some metric of utility may play, especially those which can be computed inter-datasets leading to confusion between utility and privacy evaluation, as explained in section~\ref{subsubsec:note}. First, we will start with answering why a data practitioner should always assess the privacy of the model used. Second, we will present the Trajectory User Linking (TUL) problem, applied to the specific case of synthetic trajectory datasets. Finally, we will study to what extend the TUL problem can be applied to assess the privacy of generative model. We will conclude with a presentation of a privacy attack, a Membership Inference Attack (MIA), specifically designed to work against blurring models and based on the metric in which the \textit{data preservation} category from~\citet{miranda2023sok} is grounded. 

\subsection{On the necessity to evaluate the privacy of generators}
\label{subsec:evaluateprivacy} 
Utility and privacy are dual and are linked by nature. That is why a proper evaluation of the privacy guarantees is necessary to understand the information retained by a model from its training dataset even if this goes against the shared intuition in the literature that it should be able to generalize without memorizing anything from its training dataset. There is a large amount of existing work related to this issue, compiled in multiple surveys~\cite{buchholz2024sok, monreale2023survey, kulkarni2018generative}. Furthermore, privacy evaluation should be performed through an adversarial evaluation, following the regulation, as mandated by the EU law~\cite{GDPR}, and following the literature~\cite{yao2025dcr}. 

A significant part of the literature on synthetic data generation uses distance-based metric to demonstrate their privacy preserving properties, such as distance to the closest record~\cite{lu2019empirical, sivakumar2023generativemtd, kotelnikov2023tabddpm}. As explained in the introduction,~\citet{najjar2022trajectory} demonstrate how the TUL task can be reduced to a distance base evaluation. In contrast,  recent works conducted by ~\citet{yao2025dcr} extensively studied why a distance based evaluation consistently does not reflect the real privacy risk a model presents. In their work, following the recommended adversarial view, they advise to use standard ways to assess the privacy of models, such as membership inference attacks. Recent work also explores why synthetic data generation methods do not mitigate privacy risks alone, such as in the work conducted by~\citet{ganev2026rethinking}, modifying the perception we have of synthetic solutions.  

The state of the art technique for evaluating the privacy of a tabular synthetic model is the MIA~\cite{houssiau2022tapas}. This attack infers whether an individual has been used during the training phase of a target model. As an example, one may think of an attacker trying to infer information about a model to detect cancer in Magnetic Resonance Imaging (MRI) released by a hospital. In this scenario, the hospital used their own private database of MRIs to train their model. In that case, a successful membership of a specific individual would disclose private health information about this person. In the literature, successful MIAs exist against numerous models, especially GANs~\cite{van2023membership}, and DPMs in white-box settings~\cite{pang2023white, wu2025winning} or black-box settings~\cite{wu2025winning}, all against tabular datasets. Here we follow the claim made by~\citet{yao2025dcr} and advise the data practitioner that would use one of the deep generative solutions to evaluate their model through an MIA scope. In the following sections, we will provide a novel MIA evaluation applied to synthetic trajectory dataset. 

\subsection{The TUL task, in contrast with privacy evaluation}
\label{subsec:TUL}

First introduced by~\citet{gao2017identifying}, the TUL problem can be formalized as follows: Consider a human mobility dataset $D$. On the one hand, we call $U_i$ the i-th user in $U$, with $U$ the set of all users who contributed to the human mobility dataset $D$. On the other hand, $D$ is made of trajectories defined as $T_j = [(p_{1,j},t_{1,j}),\dots,(p_{|T_j|},t_{|T_j|})]$, as presented in our background section~\ref{subsec:trajectories}. The TUL problem is to find a function $f$ that will map $D \rightarrow U$. In plain words, solving the TUL problem is to find a function $f$ that properly assigns each trajectory to its origin user. 

Before being able to apply the TUL problem, we need to distinguish between two cases. 

The first case concerns synthetic datasets, as defined in Section~\ref{subsubsec:syntheticdata}, i.e.  when the generator produces records from the underlying distribution $\mathcal{D}$ learned from its training dataset, solely using random noise as input. In this approach, there is no direct link between synthesized records and the original data during the generation stage. Therefore, any solution to the TUL problem that creates such a link would, at most, raise serious suspicions about the model generalization's capacity -- perhaps even indicating a memorization phenomenon. 

In contrast, the second case concerns generators that produce `synthetic' data by transforming or modifying existing real records sampled in $Q$, that we defined as blurring models in~\ref{subsubsec:blurringmodels}. This approach inherently creates a direct link between the generated record and its original source, where each generated record is attributable to the origin user of the blurred trajectory. The TUL task becomes a direct measure of privacy leakage in this scenario, where success allows explicit re-identification of users in $Q$ from the transformed data. This architectural characteristic, which directly impacts privacy, is observed in practical implementations, such as the one proposed by~\citet{rao2020lstm}. 

\subsection{A case study: privacy attack against blurring models}
\label{subsec:MIAblurringmodels}

We present in this section a new privacy attack against blurring models. Our motivation is to evaluate the use of TUL based methods in assessing the privacy-preserving properties of synthetic trajectory data generation algorithms. In this section, we will have a particular focus on blurring models, after we identified methodological shortcoming in previous research. To assess this privacy-preserving property, we will first conduct a review on the TUL application to produce a new methodology in the case of blurring models, in section~\ref{subsubsec:TULbased}. In the rest of the section, we will present a threshold-based MIA against blurring models to contradict privacy claims that may be be done based on a TUL experiment. 

The intuition behind our attack is to adapt the membership property from the literature to the specific use of a blurring model. 
In the context of standard synthetic trajectory data generators, as defined in the literature~\cite{shokri2017membership}, membership refers to the presence or absence of a specific target record $x_T$ in the training dataset $D_{train}$ of the target model $\mathcal{G}_{target}$.
Conversely, when considering blurring models, membership refers to the presence or absence of a target record $x_T$ in the dataset $Q_{target}$ from which the blurred dataset $S_{target}$ is generated. Importantly, since $Q_{target}$ is used as input in the generation phase, inference of membership in $Q_{target}$ plays the same role as the classic membership inference attack from the literature. In this scenario, the privacy of the algorithm is jeopardized by a successful attack on $Q_{target}$.

\subsubsection{Comments on TUL based privacy experiments}
\label{subsubsec:TULbased}

When a blurring model $\mathcal{M}$ is used to protect upon release a private dataset $Q_{target}$, an evaluation of the amount of information retained by the model is essential. A methodology for this evaluation is proposed by ~\citet{rao2020lstm} and illustrated in Figure~\ref{fig:TUL_methodology}. It proceeds by training a TUL solver with a portion of the original data $D_{train}$, and then applying it to both the dataset we want to protect, $Q_{target}$, and its blurred counterpart to be released, $S_{target}$. The degree of anonymization provided by the blurring model is evaluated by comparing the TUL solver's performance on $Q_{target}$ versus $S_{target}$.

Importantly, for the TUL solver to be properly trained and provide meaningful results within this methodology, it is imperative that each user is represented in both its training data ($D_{train}$) and the target dataset ($Q_{target}$). This requires an unbiased split of the original dataset $D$ into $D_{train}$ and $Q_{target}$, typically at a trajectory level. Consequently, the users contributing to a single trajectory must be excluded from $Q_{target}$, to prevent biasing the TUL solver, as it would inherently fail to establish links for users not represented in its training data.
 
Here, we argue that conducting such evaluation requires further methodological precision. Specifically, any TUL-based privacy assessment must be used on comparable versions of traces to avoid misleading estimations of privacy preservation. Intuitively, following the previous methodology on the TUL problem is equivalent to, at best, asking the TUL solver to find a user from a set of trajectories with modified or unknown characteristics. This is because the blurred trajectory inherits the user's assignment from the blurred model, while the blurred trajectory characteristics are new to the TUL solver. 

With this perspective, the observed decrease of TUL solver's accuracy is a foreseeable outcome,  primarily attributable to these trajectories' alterations than to a genuine gain of privacy. Consequently, from a data owner's perspective, relying on such an evaluation for blurring models can be misleading and foster a false sense of privacy. Moreover, considering an adversarial scenario, should an attacker gain access to an auxiliary dataset, it is highly plausible that the attacker's TUL solver would be trained directly on the synthetic traces themselves.

In light of these considerations, we propose specific adjustments to the evaluation of TUL solver accuracy, to ensure that valid conclusions can be drawn regarding the \textit{privacy gain} attributed to the evaluated blurring model. In our modified methodology, the data owner should first train a TUL solver on the real training dataset $D_{train}$, and then assess its performance on the real dataset used to release the information $Q_{target}$. Then, in a second step, the data owner should train a new TUL solver on the \textit{synthetic} version of the training dataset $S_{train}$, to then evaluate its performance on the blurred dataset $S_{target}$. Finally, it is the observed, potentially decreasing, performance that will help us draw conclusions about the privacy of the blurring model. Formally, if we consider the dataset $D$ to be divided into $D_{train}$ and $Q_{target}$ such that each user has at least one trajectory in $D_{train}$. Then, our suggestion is to first train a TUL solver on $D_{train}$ and test it on $Q_{target}$, and second to train a second TUL solver on $S_{train}$ and test it on $S_{target}$. The goal is then to compare the two performances of the TUL solvers. In Appendix~\ref{appendix:TUL_split} (Figure~\ref{fig:TUL_methodology}), we can see a summary of the two presented TUL methodology. 

In their paper,~\citet{rao2020lstm} used the state-of-the-art TUL solver for a dataset with semantic information: MARC~\cite{may2020marc}. They applied the MARC algorithm to the Weekly Foursquare NYC check-in dataset~\cite{may2020marc}. Finally, they followed the traditional TUL methodology, as introduced in the first paragraph of this section. In their article,~\citet{rao2020lstm} reported a significant decrease in accuracy, going from an accuracy of 94\% on the real test dataset (As shown in Figure~\ref{fig:TUL_methodology}, it is referred to as $Q_{target}$), compared to an accuracy of 46\% on the blurred version of the test dataset (referred to as $S_{target}$). In both cases, the MARC TUL solver was trained on the real train dataset (referred to as $D_{train}$). 

We compared the performance of our suggested methodology of the TUL task with the aforementioned one. We applied this methodology to the exact same architecture (LSTM-TrajGAN) and splitting methodology as introduced by~\citet{rao2020lstm}. Finally, similarly to~\citet{rao2020lstm} we used the MARC TUL solver architecture in our experiments. 

Following our methodology, we trained two different TUL solver (on $D_{train}$ and $S_{train}$) and applied it to respectively $Q_{target}$ and $S_{target}$. In the first case, we observed an accuracy of 94\%, consistent with the results presented in the paper~\cite{rao2020lstm}. In the second case, we observed an accuracy of 76\%. The observed \textit{adaptability} with this methodology is a decrease of 18 percentage points (pp), to be compared with the decrease reported in their paper and reproduced, of 48 pp. Although this still gives some privacy gain over the real test dataset, it somehow mitigates the degree of anonymization proposed by LSTM-TrajGAN .

\subsubsection{A membership inference attack against blurring models.}
\label{subsubssec:methodology}

After having demonstrated the need for a methodological adjustment in the application of TUL task, when this is applicable, we will now investigate a more standard privacy attack, as advised by~\citet{yao2025dcr}. We will present a new, to the best of our knowledge, membership inference attack against $Q_{target}$, the dataset that the data owner aims to safeguard with the blurring model. 
This methodology focuses on the case where $D$ is made \textit{only} of spatial trajectories for brevity. The case where $D$ includes additional semantic information leads to the same conclusions and is also discussed in the results section. The intuition behind the attack is that if the models are memorizing rather than generalizing over the training datasets, giving access to data drawn from the same underlying distribution can be exploited to derive a threshold over distance metric computed between a target and members (or non-members) of the training dataset.

\textbf{Methodology overview.} We consider a dataset of human mobility $D$ with an underlying distribution $\mathcal{D}$. To respect the split made to train the blurring models, we split the dataset $D$ into $D_{train}$ and $Q_{target}$, where $Q_{target}$ will be used only to generate the blurred information. $D_{train}$ will be used to train a blurring model $\mathcal{G}_{target}$, and we denote the attacker by $\mathcal{A}$. The purpose of this attack is to infer if a specific trace $x_T$ has been used to generate the blurred dataset $S_{target} = \mathcal{G}_{target}(Q_{target})$. Our attack is based on the intuition that blurred records, in $S_{target}$, will yield closer `synthetic' counterpart than records that are not used.

To learn a threshold between members and non-members, we first consider the case where $\mathcal{A}$ has access to the following information: 
\begin{itemize}
    \item An auxiliary dataset $D_{aux}$, that follows the same underlying distribution $\mathcal{D}$ ($D_{aux} \sim \mathcal{D}$).
    \item The target record $x_T$.
    \item The target model architecture and hyperparameters. 
\end{itemize}
From it, $\mathcal{A}$ will first split $D_{aux}$ into three different datasets, $D_{aux_{train}}$, $Q_{aux}$ and $Q_{\tau}$. $D_{aux_{train}}$ will be used to train an auxiliary model using the same architecture as the target and $Q_{aux}$ will be used to generate the blurred records $S_{aux}$. Lastly, using records sampled uniformly from $Q_{aux}\cup Q_{\tau}$, $\mathcal{A}$ will be able to infer a threshold $\tau$ to distinguish between members and non-members of $Q_{aux}$. By applying this threshold to the target blurred dataset, $\mathcal{A}$ will then be able to conclude the attack. A representation of the different datasets is shown in Appendix~\ref{appendix:data_division} (Figure~\ref{fig:dataset_splits}).

\begin{figure}[!ht]    
    \centering
    \includegraphics[width=0.60\linewidth]{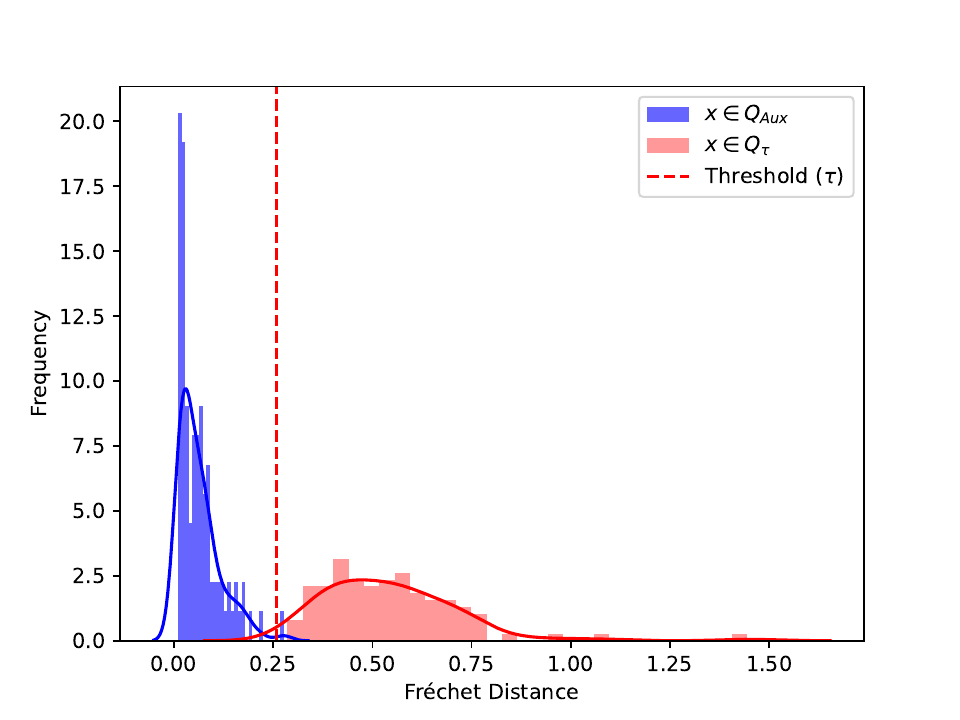}
    \caption{Example of threshold computation, based on the Fréchet distance~\cite{frechet1906quelques} and the choice of threshold formula presented in Section~\ref{subsubsec:results}.}
    \label{fig:threshold}
\end{figure}

\label{subsubsec_bold:threshold}
\textbf{Threshold computation.} After $\mathcal{A}$ split $D_{aux}$ into three parts $D_{aux_{train}}$, $Q_{aux}$ and $Q_{\tau}$. The attacker then trains a shadow model $\mathcal{G}_{aux}$ using $D_{aux_{train}}$, following the exact same architecture and hyper-parameters $\mathcal{G}_{target}$. 

\begin{table*}[htbp!]
    \centering
    \begin{tabular}{cccccccc}
        \toprule
        \multirow{3.5}{*}{\centering Dataset} & \multirow{3.5}{*}{\centering Models} & \multicolumn{6}{c}{Accuracy results} \\
         & & \multicolumn{3}{c}{Fréchet} & \multicolumn{3}{c}{Custom} \\
         \cmidrule(r){3-5} \cmidrule(r){6-8}
         & & Main & Masked & Released & Main & Masked & Released \\
         \midrule
         Weekly FS-NYC & \multicolumn{1}{l}{LSTM-TrajGAN~\cite{rao2020lstm}} & 0.883 $\pm$ 0.013 & 0.838 $\pm$ 0.019 & 0.925 $\pm$ 0.011 & 0.985 $\pm$ 0.011 & 0.818 $\pm$ 0.022 & 0.985 $\pm$ 0.015 \\
         \cite{rao2020lstm}& \multicolumn{1}{l}{exGAN~\cite{song2023except}} & 0.740 $\pm$ 0.045 & 0.740 $\pm$ 0.010 & 0.705 $\pm$ 0.043 & 0.800 $\pm$ 0.165 & 0.660 $\pm$ 0.082 & 0.700 $\pm$ 0.142 \\
         \midrule
         Geolife Daily & \multicolumn{1}{l}{LSTM-TrajGAN~\cite{rao2020lstm}} & 0.833 $\pm$ 0.048 & 0.793 $\pm$ 0.033 & 0.870 $\pm$ 0.041 & 0.945 $\pm$ 0.025 & 0.790 $\pm$ 0.031 & 0.950 $\pm$ 0.023 \\
         \cite{zheng2008understanding, zheng2009mining, zheng2010geolife} & \multicolumn{1}{l}{exGAN~\cite{song2023except}} & 0.833 $\pm$ 0.004 & 0.808 $\pm$ 0.027 & 0.775 $\pm$ 0.044 & 0.915 $\pm$ 0.029 & 0.765 $\pm$ 0.023 & 0.933 $\pm$ 0.026 \\
         \bottomrule
    \end{tabular}
    \caption{Results of our attacks. Results are presented over the two datasets used in both use-cases. Results are derived over the two different metric used to compute the distance (Fréchet and Custom) and the three different scenarios considered, presented using the form (Mean $\pm$ std).}
    \label{tab:table_mia_results}
\end{table*}

To infer the threshold allowing a separation between members and non-members, the attacker generates synthetic records from $\mathcal{G}_{aux}$, using $Q_{aux}$. Formally, $\mathcal{A}$ generates $S_{aux} = \mathcal{G}_{aux}(Q_{aux})$. We refer to the distance between two trajectories $T_i$ and $T_j$ as $d(T_i,T_j)$. With a slight abuse of notation, we refer as $\overline{d(A,B)}$ as the mean distance between all the points in $A$ and $B$. 

Once this computation is completed, the attacker can infer the threshold $\tau$. This threshold depends on the choice of distance $d$ and the choice of relation between the two distributions $d(Q_{\tau}, S_{aux})$ and $d(Q_{aux}, S_{aux})$. To be considered a suitable threshold, $\tau$ need to separate the two distribution of the auxiliary data convincingly.
An example of threshold computation, based on the choices we made in the next section, can be found in Figure~\ref{fig:threshold}.

\textbf{Membership Inference.} After computing the threshold $\tau$, $\mathcal{A}$ only has to apply it to the blurred dataset $S_{target}$. In practice, $\mathcal{A}$ takes $x_T$ and computes $\alpha = \text{min}(\{d(x_T, y) ~|~\forall y \in S_{target}\})$. 

This score $\alpha$ is attributed to $x_T$, and $\mathcal{A}$ infers the membership of $x_T$ in $Q_{target}$ based on its value. Specifically, $\mathcal{A}$ predicts `IN' if $\alpha \leq \tau$, `OUT' otherwise. 

These results depend heavily on the amount of information we give the attacker access to. We provide two different variants to restrict this knowledge, targeting two different sources of information.
\subsubsection{Variants.}
Here we present variants of the attack to relax different assumptions made on the knowledge of the attacker.

\textbf{Variant 1: Masked.} In practice, giving full knowledge of the target to the attacker may be a strong assumption. Here we suggest applying the previous methodology to a masked target, when only a fraction of the real trace is given to the attacker. In this scenario, the attacker has access to only a fraction $\beta$ of the target trace $x_T$. This mask application does not modify the structure of the attack. It only impacts the precision behind the computation of any distance, giving more uncertainty on the computation of the threshold and its application. 

\textbf{Variant 2: Released only.} In this second variant, we consider a weaker attacker (e.g., more realistic) who does not have access to the auxiliary dataset $D_{aux}$, but only to the blurred dataset released, $S_{target}$. The intuition behind this variant is that the blurred dataset delivered by the target still contains all the information needed by an attacker to carry out the attack. 

First, the attacker performs on $S_{target}$, the same split as was originally done with $D_{aux}$: $S_{target}$ is split into three different datasets, $SD_{aux}$, $SQ_{aux}$ and $SQ_{\tau}$, such that $S_{target} = SD_{aux}\cup SQ_{aux}\cup SQ_{\tau}$. The attacker then proceeds as described in~\ref{subsubssec:methodology}, training a shadow model on $SD_{aux}$, and inferring the threshold based on $SQ_{aux}$ and $SQ_{\tau}$. The application of the threshold on the target's score is not modified by this variant of the attack. 

\subsubsection{Results.}
\label{subsubsec:results}
By applying the previous methodology on multiple targets, we can compute the performances of this attack. In this paper, we made the following choices to compute the results.

\textbf{(1)} We choose two different metrics. The first one, referred as `Fréchet' in Table~\ref{tab:table_mia_results}, only considers spatial trajectories. To apply this metric, we considered only the trajectories with the same number of visits as the target $x_T$ in the dataset released. The second one considered in this paper is named \textit{custom}, and is the sum of the Fréchet distance and the cosine similarities between all other semantic features. Similarly, we only consider the trajectories with the same number of visits as the target $x_T$.

\textbf{(2)} We choose all blurring models among the three models used in the previous section~\ref{sec:utility}. We could not apply our MIA to the last model, TrajGDM~\cite{chu2024simulating}, as this is not a blurred model. Since both~\citet{rao2020lstm} and~\citet{song2023except} used the TUL problem to justify the privacy-preserving properties of LSTM-TrajGAN (respectively exGAN) in their original articles.

\textbf{(3)} We consider as a threshold the midpoint between the two means of the distributions. We arbitrarily choose the center of mass here, and left for future work the exploration of better thresholds. Formally, this can be written as follows: \newline $\tau = \frac{\overline{d(Q_{\tau}, S_{aux})} + \overline{d(Q_{aux}, S_{aux})}}{2}$ with $d$ being one of the two previous metrics.
    
\textbf{(4)} To compare our results, we choose to consider only accuracy.~\citet{rao2020lstm} used the Weekly Foursquare NYC check-ins dataset (from~\cite{may2020marc}), so to allow a fair comparison, we here computed results on the exact same dataset to compare our performances. For completeness with our utility study, we also trained our models over the Geolife~\cite{zheng2008understanding,zheng2009mining,zheng2010geolife} dataset and reported the results over the two different models. Results are computed over 4 different seeds.

First, as we can see in Table~\ref{tab:table_mia_results}, our attack achieves great performance and is able to infer membership in all configurations with great certainty, up to 98.5 $\pm$ 1.1 \% accuracy for our custom metric. When we consider LSTM-TrajGAN models, trained on both datasets in the main scenario, computed with the Fréchet metric, the membership inference attack achieves a score of 88.3 $\pm$ 1.3 \% and 83.3 $\pm$ 4.8 \% accuracy respectively. Even greater, when we consider the custom metric is the exact same scenarios, this score goes up to 98.5 $\pm$ 1.1\% and 94.5 $\pm$ 2.5 \% accuracy. Similarly, for exGAN, in the main scenario, the membership inference attack achieved respectively a score of 74.0 $\pm$ 4.5 \% and 83.3 $\pm$ 0.4\% using the Fréchet metric, and respectively 80.0 $\pm$ 16.5 \% and 91.5 $\pm$ 2.9 \% with our custom metric. In practice, it means that the use of this model to release information will disclose the membership of any member, with a slightly better protection offered by exGAN. If we observe the consequence of such an attack in the context of our hospital example, it means releasing information after blurring it with LSTM-TrajGAN would result in a potential disclosure of patients' health information. A successful membership against $x_T$ would result in the conclusion $x_T$ was ill at the time of the generation. 

In all our privacy experiments, we used a constant number of epochs, the same than originally used by the paper introducing the models. Note that using more epochs to train the model would improve the attack results, especially for exGAN. A study of the number of epochs necessary to train the models can be found in Appendix~\ref{appendix:epochs}.

We also explored different variant scenarios where we weakened the information given to the adversary, making attack scenarios more realistic. In our first variant, called \textit{Masked}, We fixed a mask level of 75\%, which means that only 25\% of the visits of the target $x_T$ are given to the attacker $\mathcal{A}$. In this variant scenario, our best attack configuration reported an accuracy of 83..8 $\pm$ 1.9 \% for and LSTM-TrajGAN trained with the weekly FS-NYC dataset. With every configuration achieving great accuracy score, it means that even with partial knowledge of the $x_T$ trajectory, such as the work place and the house information from our target $x_T$, despite this lack of information the attack's performances would remain excellent. Results of our custom metric in the exact same attack configuration decreased slightly to 81.8 $\pm$ 2.2 \%, also observed for other attack configurations. This can be explained because in our custom metric, we balanced the importance of each feature. The conclusion is that with a masked target $x_T$, this balance perturbs the transferability of our threshold, probably because it lacks information. 

In our second variant, called $\textit{Synthetic Only}$, we now only give access to the released information to the attacker (eg., the released synthetic dataset). In this particular setting, we significantly decrease the information given by the attacker, as the attack is now based solely on what is explicitly given to the public. Surprisingly, we did not observe an expected significant decrease between the main attack scenario and this variant. Instead, we observed similar results between the two. We either increase the results, as for the LSTM-TrajGAN trained on Weekly FS-NYC with the Fréchet metric, where we increase the accuracy by 4.2\% on average, or decrease it, as for the exGAN trained on Geolife Daily, still with the Fréchet metric, where we decrease the accuracy by 5.8 \% in average. This comparable performances finish to demonstrate the inherent vulnerability of the dataset generated using both models architecture, in every single configuration. Indeed, the use of the released information does not have an impact on the attack performances, which means the released dataset and the real dataset contain the almost same level of private information. This result, aligned with previous studies~\cite{guepin2023synthetic}, grounded our conclusion on the privacy of blurring models. 

Our privacy analysis demonstrates that blurring models should not be deployed in practice without rigorous privacy evaluation when confidentiality is a priority. In cases where privacy concerns are secondary, the application of such models should be restricted to specific internal use by the data owner, such as data augmentation for learning tasks where additional samples are required. With our two use-cases, it means the data owners also need to define the level of privacy required by their applications, as it largely impacts the choice for the model. In every scenario of attack we developed, even when the attacker only has access to the information released, our results show that a successful membership inference attack against blurring models can be conducted. Following those results, we hence suggest not to use the TUL problem to assess the privacy of a model, even with our methodology. Our attack demonstrated the intuition we can derive from the literature: Although the TUL problem is often seen as a difficult task, one that would need large models to be solved,~\citet{najjar2022trajectory} showed how the TUL problem is easier than previously thought. In fact, trajectory data are very specific and very little information is needed to allow some privacy risks. Similarly to the work by~\citet{de2013unique},~\citet{najjar2022trajectory} showed that a trajectory is highly unique, leading to the development of a heuristic-based TUL solver. This proof tends to show that the TUL evaluation can be reduced to a distance-based privacy evaluation, when applicable. Together, with the work by~\citet{yao2025dcr}, it built the intuition that TUL solver tasks could give a false sense of privacy if used to assess it.

Finally, as highlighted by~\citet{yao2025dcr} and~\citet{houssiau2022tapas} (and further detailed in Section~\ref{subsec:evaluateprivacy}), while various attacks exist to assess a model's privacy properties, membership inference attacks remain the established state-of-the-art approach for privacy evaluation. Consequently, they should be considered the primary tool in most data privacy assessment paradigms, as we did in our Section~\ref{subsec:evaluateprivacy}.

%% file: related_work.tex
\textbf{Utility evaluation for synthetic dataset of mobility.} The need for a proper evaluation of the performance of generative models has been identified since their beginnings. In these pioneering works ~\citet{liu2018trajgans} establish distinctions between several use cases, such as speed distribution and network-based urban structure, emphasizing their differences from an utility perspective. However, they did not provide a comprehensive segmentation of utility evaluation based on this distinction. Since then, this challenge has been addressed by articles that introduced new trajectory generation models~\cite{rao2020lstm,feng2020learning,chen2021trajvae,zhu2023difftraj,cao2021generating}. These varied works present different aspects of utility evaluation, but they are hard to compare since each paper used different metric of utility, often covering orthogonal aspects of utility. Following this paper,~\citet{miranda2023sok} in a survey about how private the mechanisms are, introduce three key axis of utility evaluation. Namely, data preservation, statistics preservation and realism assurance. While these distinctions are crucial, they do not encompass the full complexity  and specificities of trajectories datasets. Inspired by the standard practices for evaluating utility in the generation of synthetic tabular dataset~\cite{bowen2021utility, mckenna2021winning}, we think that having a single column to represent the statistical information contained in the dataset is not enough to grasp its global complexity. Our work introduces a distinction between marginal and relational statistics and presents a new category called `Task Performance' to further understand how the traces are used on the downstream tasks. A recent survey by ~\citet{buchholz2024sok} has introduced two levels of utility evaluation, through a first attempt of utility metric systematization (in their Table 1) inspired by the original distinction made by ~\citet{liu2018trajgans}. In their survey, they present two different categories along two different axes: the statistical information from the data, and their realism are described along two different focus, described as point-level and trajectory-level. We think this systematization could benefit from the following categories: The ones introduced by~\citet{miranda2023sok}, and the ones introduced before. In our work, we build upon these two surveys to introduce a systematization of utility evaluation through the scope of four distinct aspects of utility and two different levels of information. 

\textbf{Privacy evaluation of synthetic tabular dataset.} An extensive literature exists on the privacy evaluation of synthetic tabular dataset, that can be separated in four main categories.
First, membership inference attacks~\cite{shokri2017membership, carlini2022membership, salem2018ml, yeom2018privacy}, where an attacker tries to infer whether or not a record has been used during the training of a target model. Second, attribute inference attacks~\cite{fredrikson2014privacy, fredrikson2015model, mehnaz2022your, annamalai2024linear}, where an attacker tries to infer private information about a record knowing the model and the publicly available information about this individual. Third, property inference attacks~\cite{ateniese2015hacking, ganju2018property} where an attacker tries to infer private statistical information about the training dataset of a target model, and lastly reconstruction attacks~\cite{balle2022reconstructing, salem2020updates}, where the attacker tries to take advantage of the model to reconstruct the information of training data samples. 
None of those standard privacy evaluations, to the best of our knowledge, has been previously adapted and used in trajectory generation. In our work we introduced the need for careful privacy evaluation, by showing how a membership inference attack could reveal important private information. 

%% file: conclusion.tex
In this paper, we proposed a method for a systematic evaluation of utility across models. We provide an applicable framework for utility evaluation. This framework allows for a systematic evaluation of the utility across datasets and deep generative models. To show how this framework can be used to compare models, we developed three different examples of utility computation and compared them. Being able to compare utility coherently opens the path to solving the privacy-utility trade-off in mobility data generation. 

The next step toward solving this trade-off is the privacy evaluation. Our study highlighted limitations in current privacy evaluation practices. To explain these limitations, we introduced a distinction between synthetic and blurring models. On the one hand, models produce data from sampling in the learned distribution, on the other hand, models blur the real information to protect it. For blurring models we showed that sole reliance on the TUL task as a privacy indicator can be misleading. We further demonstrated that such models, believed to be private due to a significant decrease in TUL score, are still vulnerable to privacy leakage. To demonstrate this, we developed a successful MIA that we applied to two different blurring models, over two different real datasets. For synthetic models we identified that most evaluations relied only on their randomness. Grounding our analysis in state-of-the-art privacy evaluations, we argued in favor of a systematic privacy evaluation framework. We demonstrated that the evaluation of privacy in generative models should be carefully considered. While we focused our paper on MIAs, a framework of privacy evaluation should also take into consideration, with careful adaptation, the landscape of privacy attacks, such as reconstruction, correlation and inference attacks. Furthermore, we purposely did not investigate models that included differential privacy in their architecture. Differential privacy refers to the addition of controlled noise to the training process of a mechanism, to avoid being able to single out any member of the training dataset. Addition of differential privacy is providing for privacy guarantee in models. While we believe our framework to remain applicable, it would have been out of scope for the privacy discussion presented in this work. 

\textbf{Future Work.} We left for future work improvements over the different choices in the definition of distance metric we made in the membership inference attack. We left for future work to develop a similar systematic approach for the privacy evaluation of generative models. Ultimately, we left for future work the articulation between the two frameworks (utility and privacy), leading to careful consideration about the privacy-utility trade-off. As a final step, we leave for future work the inclusion of protected generative models (e.g., differentially private models) in the given frameworks.

%% file: appendix.tex
\begin{table}[htbp!]
    \centering
    \begin{tabular}{cl}
    \toprule
        $D$ & A human mobility dataset \\
        $\mathcal{D}$ & The underlying distribution of a dataset $D$\\
        $U$ & The set of all users contributing to a dataset $D$\\
        $T$ & A trajectory \\
        $\Omega$ & The space of all possible trajectories\\
        $\Phi_D$ & A generator trained on the dataset $D$ \\
        $\tilde{\Phi}_D$ & A blurring model trained on the dataset $D$\\
        $D_{syn}$ & A synthetic dataset generated by $\Phi_D$ or $\tilde{\Phi}_D$\\
        $D_{train}$ & The set used to train a target model \\
        $Q_{target}$ & The set used to test a target model \\
        $S_Y$ & The synthetic/blurred version of any dataset $X_Y$ \\
        $\mathcal{A}$ & The attacker \\
        $x_T$ & The target trajectory \\
        $D_{aux}$ & The auxiliary information available to the attacker\\
        $\tau$ & A threshold used and introduced by $\mathcal{A}$ in our attack\\
    \bottomrule
    \end{tabular}
    \caption{Notation glossary}
    \label{tab:notations}
\end{table}
\section{Notation Glossary}
\label{appendix:glossary}
Table~\ref{tab:notations} provides a reminder of all the notation used throughout the paper. 
\section{TUL methodology}
\label{appendix:TUL_split}
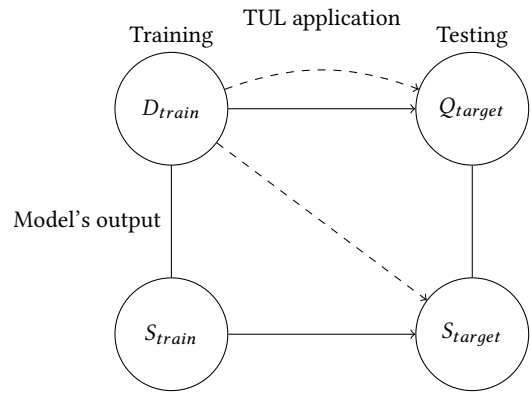
\begin{figure}[htbp!]
    \centering
    \begin{tikzpicture}[node distance=4cm, every node/.style={circle, draw, minimum size=1.5cm}]

      \node (1) {$D_{train}$};
      \node (2) [right of=1] {$Q_{target}$};
      \node (3) [below = -1.0cm, below of=1] {$S_{train}$};
      \node (4) [below = -1.0cm, below of=2] {$S_{target}$};
      \node (5) [above=0.2cm, draw=none, fill=none, rectangle, inner sep=0pt, outer sep=0pt] {Training};
      \node (6) [above=-0.55cm of 2, draw=none, fill=none, rectangle, inner sep=0pt, outer sep=0pt] {Testing};

      \draw[dashed, ->, bend left=20] (1) to (2);
      \draw[dashed, ->] (1) -- (4);

      \draw[->] (1) -- (2) node[midway, above, draw=none] {\textcolor{black}{TUL application}};
      \draw[->] (3) -- (4);

      \draw[-] (1) -- (3) node[midway, left, draw=none] {Model's output};
      \draw[-] (2) -- (4) ;

    \end{tikzpicture}
    \caption{Presentation of the two different methodology for the TUL application. In dash line, it represents the traditional TUL application, using a single TUL solver trained on $D_{train}$ as presented in~\citet{rao2020lstm}, while in plain line it present our suggestion, training two different TUL solvers on both $D_{train}$ and $S_{train}$ to compare their performances.}
    \label{fig:TUL_methodology}
\end{figure}

Figure~\ref{fig:TUL_methodology} presents the previous TUL methodology (in dashed lines) and our fix suggestion (in plain lines). TUL problem is used in this methodology as a mean to evaluate the privacy level of a blurring model. 

\section{Dataset Division}
\label{appendix:data_division}
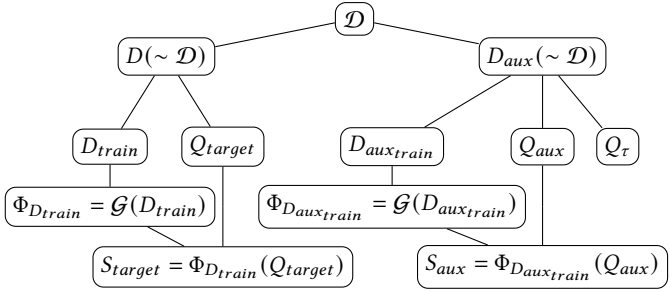
\begin{figure}[!htbp]
    \centering
    \begin{tikzpicture}[
    level 1/.style={sibling distance=50mm, level distance = 5mm},
    level 2/.style={sibling distance=15mm, level distance = 12mm},
    level 3/.style={sibling distance=1mm, level distance = 8mm},
    level 4/.style={sibling distance=1mm},
    every node/.style={draw, rounded corners, font=\sffamily}
    ]
      \node {$\mathcal{D}$}
        child { node {$D (\sim \mathcal{D})$} 
          child { node {$D_{train}$}
            child { node (nodeA) {$\Phi_{D_{train}} = \mathcal{G}(D_{train})$}}
          }
          child { node {$Q_{target}$} 
                child[level distance=8mm] {
                    child { node (nodeB) {$S_{target} = \Phi_{D_{train}}(Q_{target})$}}
                }
            }
        }
        child { node {$D_{aux} (\sim \mathcal{D})$}
          child[sibling distance=20mm] { node {$D_{aux_{train}}$} 
            child { node (nodeC) {$\Phi_{D_{aux_{train}}} = \mathcal{G}(D_{aux_{train}})$}}
          }
          child { node {$Q_{aux}$} 
              child[level distance=8mm] { 
                child {node (nodeD) {$S_{aux} = \Phi_{D_{aux_{train}}}(Q_{aux})$}}
              }
            }
          child[sibling distance=10mm] { node {$Q_{\tau}$} }
        };
        \draw (nodeA) to (nodeB);
        \draw (nodeC) to (nodeD);
    \end{tikzpicture}
    \caption{Representation of the different datasets, models (with the fitting procedure $\mathcal{G}$), and underlying distribution ($\mathcal{D}$)}
    \label{fig:dataset_splits}
\end{figure}

Figure~\ref{fig:dataset_splits} explains how the division is done to access all the key datasets used in the attack detailed in Section~\ref{subsec:MIAblurringmodels}. $\Phi_X$ denotes the deep blurring models obtained by applying a fitting procedure $\mathcal{G}$ known by the attacker to a specific training dataset $X$. Then, $\Phi_X$ is applied to the dataset $Q_Y$ to obtain its blurred counterpart, $S_Y$.

\section{Training setup}
\label{appendix:training_setup}
\subsection{Models hyperparameters}
\label{appendix:hyperparameters}
We report in Table~\ref{tab:hyperparam} the optimization hyperparameters used for training the generative models reported in Section~\ref{sec:utility}. The full set of hyperparameters is also available in the accompanying code repository\footnote{https://github.com/annonymizedanonym/annonymouspaper}.
\begin{table*}[htbp!]
    \centering
    \begin{tabular}{ccccccc}
    \toprule
    \textbf{Model} & \textbf{Batch size} & \textbf{Optimizer} & \textbf{LR} & $\beta_1$ &$\beta_2$ &  \textbf{Loss} \\
    \midrule
         \textbf{LSTM-TrajGAN}  & 256 & Adam & 0.001 & 0.5& 0.999 & TrajLoss\\
         \textbf{exGAN}  & 256 & Adam & 0.001 & 0.5 & 0.999 & TrajLoss\\
         \textbf{TrajGDM} & 64 & Adam & 0.0001 & 0.9 & 0.999 & SoftMax Cross-Entropy\\
    \bottomrule
    \end{tabular}
    \caption{Training hyperparameters for all generative models evaluated in this work. LR denotes the learning rate. $\beta_1$ and $\beta_2$ are the exponential decay rates for the 1st and 2nd moment estimates in the Adam optimizer, respectively.}
    \label{tab:hyperparam}
\end{table*}
To accommodate the inherent variability in the sampling rate and trip durations within the weekly Foursquare NYC and the three splits of the Geolife dataset (weekly, daily, hourly), we extended the TrajGDM architecture to support variable-length sequences via zero-padding and dynamic masking. Our implementation reproduces the original model’s performance on its reference dataset~\cite{chu2024simulating}. As for LSTM-TrajGAN and exGAN, we maintained the same architecture as the original paper, as they support variable-length sequences.
\subsection{Data preprocessing}
To ensure comparability between models, we adopted the same spatial normalization strategy for both the LSTM-TrajGAN and its variant exGAN. Specifically, coordinates were centered using the global centroid of the entire dataset. While we acknowledge this introduces a degree of data leakage, it was maintained to preserve the spatial encoding characteristics intended by the original authors. This ensures that our implementation and benchmark are measured against the baseline's exact operating condition.
For TrajGDM, we used a row-major spatial indexation with a 500 meter cell edge length for all datasets as it was done in the original work.
\subsection{Model convergence}
\label{appendix:epochs}
\begin{figure*}[htbp]
    \centering
    \begin{subfigure}[b]{0.32\textwidth}
        \centering
        \includegraphics[width=\textwidth]{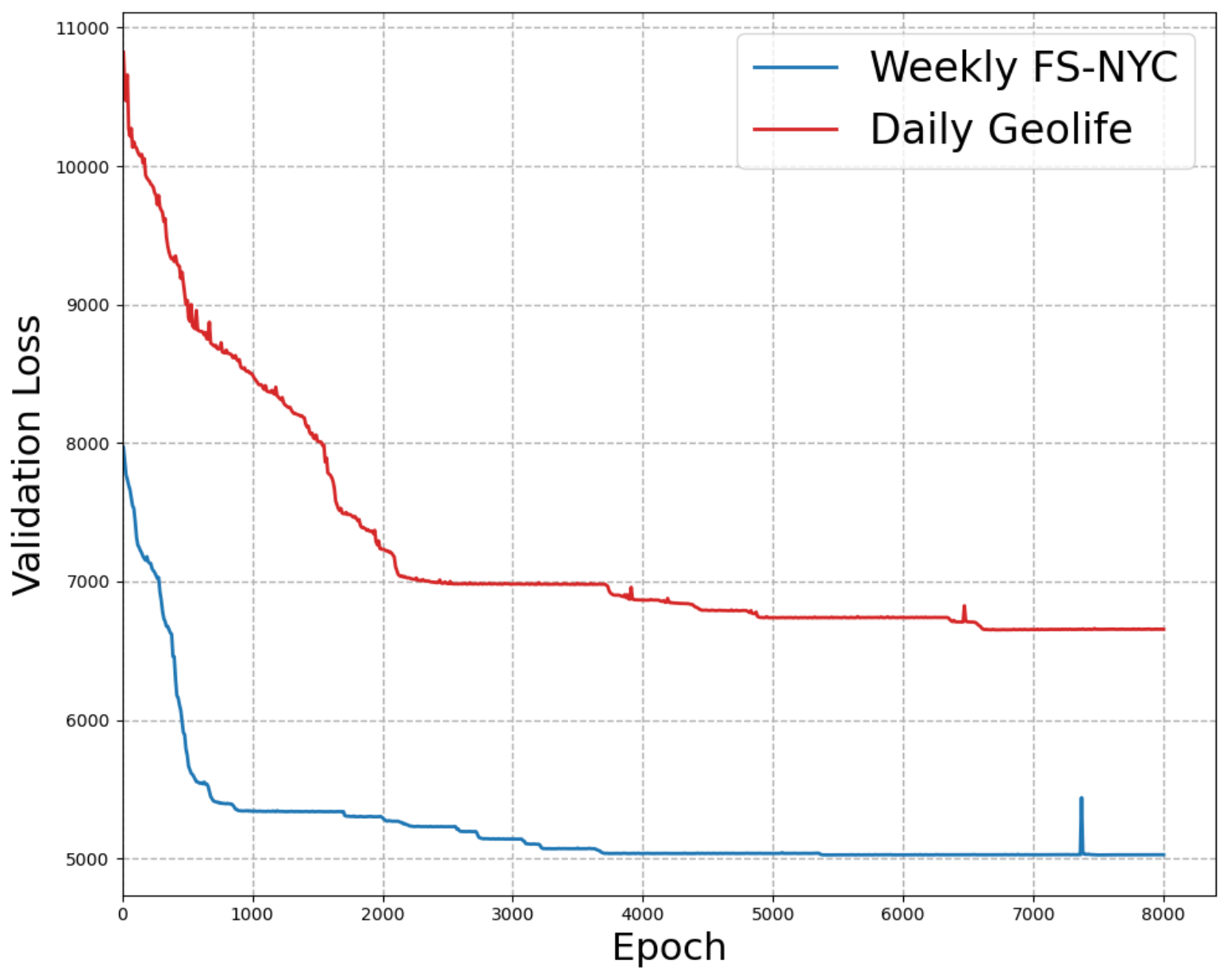}
        \caption{Generator's loss of LSTM-TrajGAN}
        \label{fig:subfig_b}
    \end{subfigure}
    \hfill
    \begin{subfigure}[b]{0.32\textwidth}
        \centering
        \includegraphics[width=\textwidth]{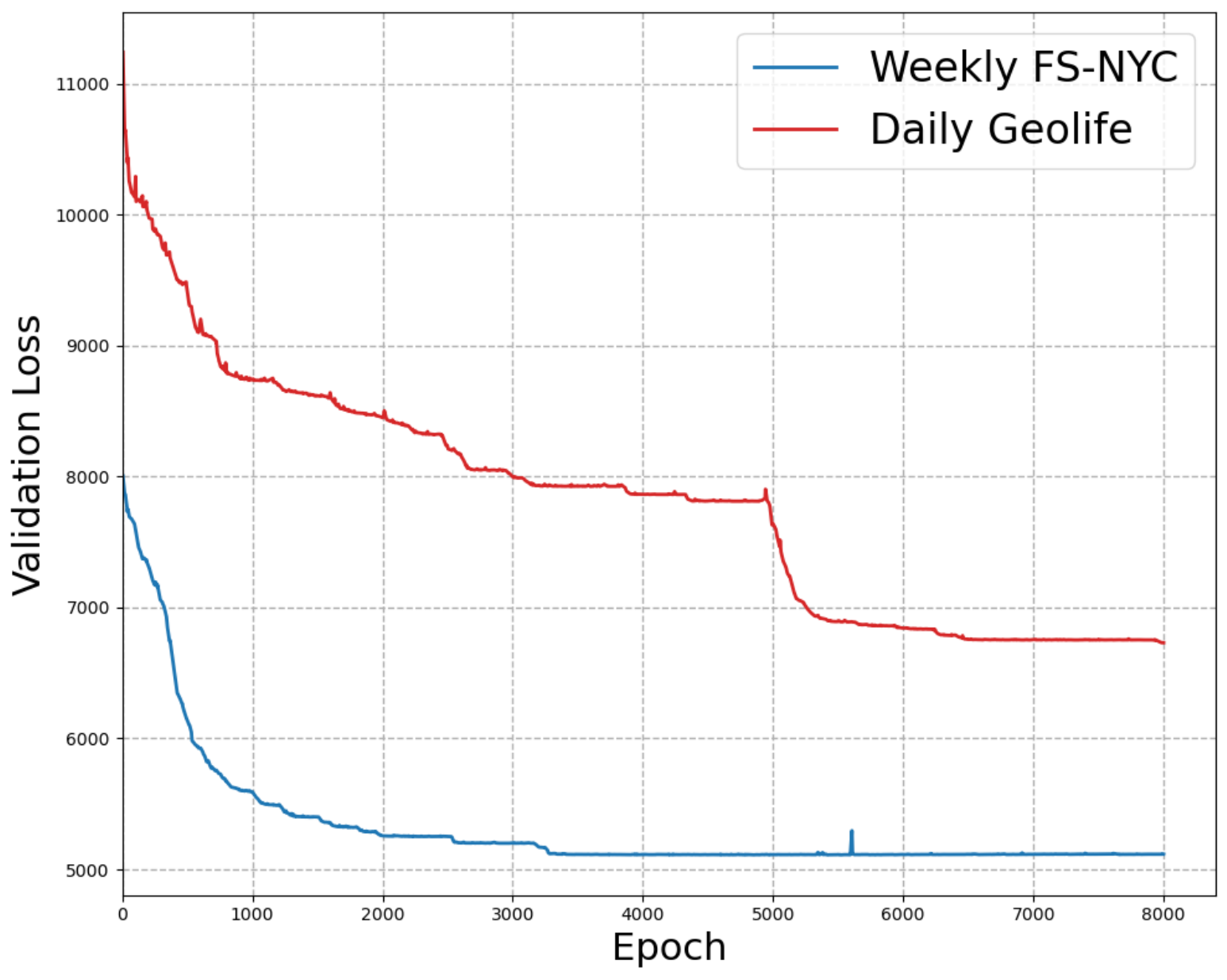}
        \caption{Generator's loss of exGAN}
        \label{fig:subfig_a}
    \end{subfigure}
    \hfill
    \begin{subfigure}[b]{0.32\textwidth}
        \centering
        \includegraphics[width=\textwidth]{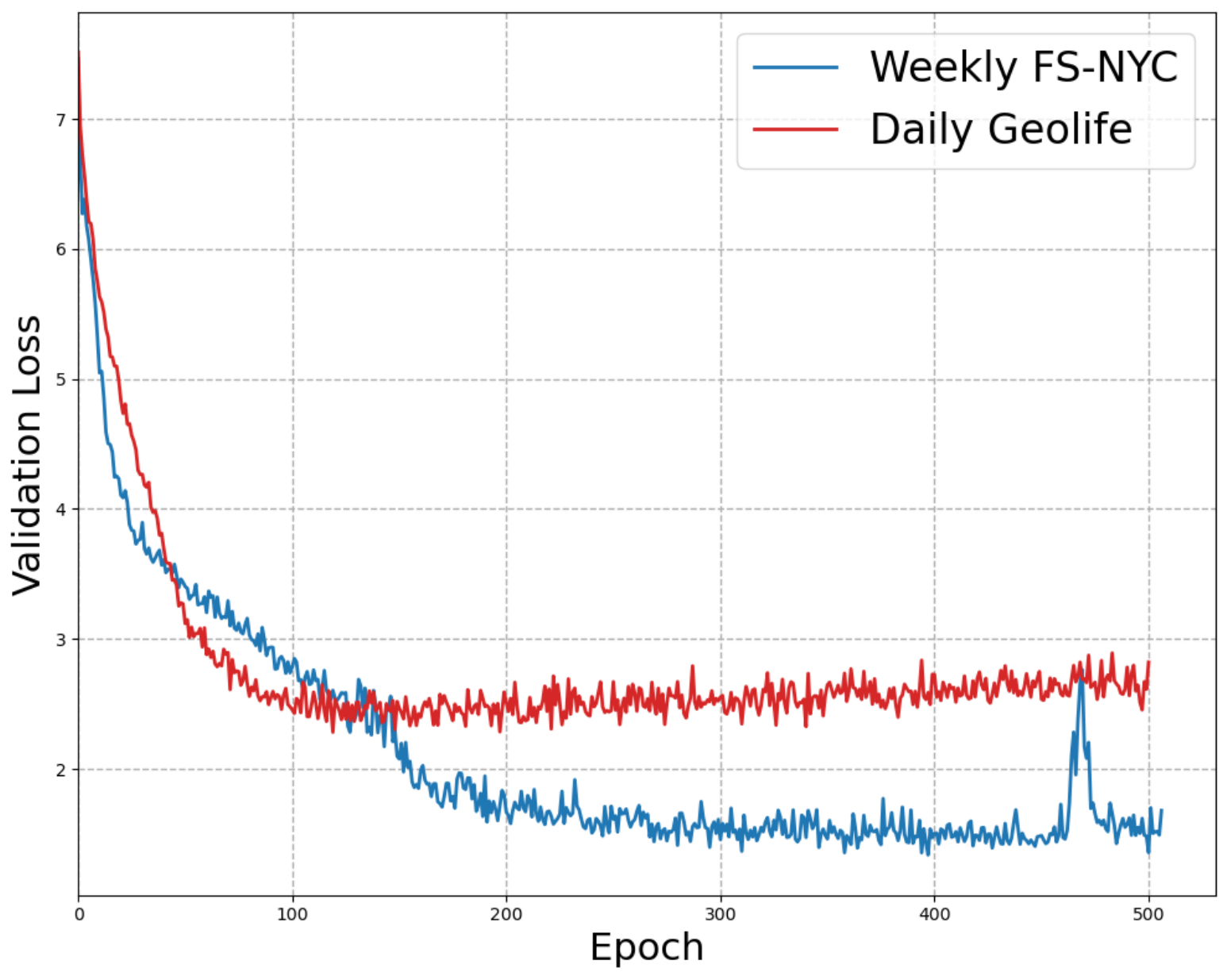}
        \caption{Cross entropy loss of TrajGDM}
        \label{fig:subfig_c}
    \end{subfigure}
    \caption{Validation loss curves for LSTM-TrajGAN, exGAN and TrajGDM models}
    \label{fig:convergence}
\end{figure*}

In this section, we report the convergence profile of each of our three models. Figure~\ref{fig:convergence} show the generator's loss for LSTM-TrajGAN and exGAN and the loss for trajGDM on validation sets for the datasets weekly FS-NYC and daily geolife.
\section{Utility Evaluation Metrics}
\label{appendix:utilitymetrics}
In this section, we present the utility evaluation metrics and evaluation protocol used in tables~\ref{tab:usecaseA} and~\ref{tab:usecaseB}.
Note that all comparisons between the original dataset (test set) and its synthetic counterpart at the trajectory-level, do not involve any one-to-one mapping, as per our methodological note~\ref{subsubsec:note}.

\subsection{Statistics preservation}
The evaluation of the preservation of statistics involves two steps: first the extraction of the statistics in question from the original and synthetic dataset separately, secondly we compare the resulting statistics, in the form of a distribution using a distributional distance. In this subsection we first introduce the measures we use to compare resulting distributions, then we introduce the statistics we included and how we compared them.
\subsubsection{Distributional distances}

\paragraph{Wasserstein Distance~\cite{santambrogio2015optimal}}
\label{appendix:def_wasserstein}
We quantify the discrepancy between distributions $\mu$ and $\nu$ using the 1-Wasserstein distance also known as the Earth Mover's Distance~\cite{rubner2000earth}. Intuitively, the 1-Wasserstein distance represents the minimum \textit{work} required to transform one distribution into another, where work is defined as the product of the amount of mass moved and the distance of transport.

Formally, let $\mu = \sum_{i} a_i \delta_{x_i}$ and $\nu = \sum_{j} b_j \delta_{y_j}$ be two discrete distributions with weight vectors $a$ and $b$, the 1-Wasserstein distance, $W_1$ is defined as the solution to:
\begin{equation}
W_1(\mu, \nu) = \min_{\gamma \in \Pi(a, b)} \sum_{i,j} \gamma_{ij} C_{ij}
\label{eq:1w}
\end{equation}

The optimization yields an optimal transport plan $\gamma \in \mathbb{R}_+^{n \times m}$, where each entry $\gamma_{ij}$ represents the amount of mass transported from element $x_i$ to $y_j$ in the support $\mathcal{X}$. The set of feasible plans is constrained by the transport poly-tope $\Pi(a, b) = \{ \gamma \in \mathbb{R}_+^{n \times m} \mid \gamma \mathbf{1}_m = a, \gamma^T \mathbf{1}_n = b \}$, ensuring mass conservation. The $C_{ij}$ represents the distance between $x_i$ and $y_j$.

We employ two distinct specifications for $C$ based on the object being compared.
For non-spatial distributions (e.g., distributions of per-trajectory average speeds, distributions of pairwise distance), we define the cost as $C_{ij}= \|x_i-y_j\|_2$. In this case $W_1$ retains the same physical units as the scalar (e.g., km/h).

For distributions defined over discrete grid cells (e.g., transition probabilities), we incorporates the physical geometry of the domain. We define the spatial ground cost $C_{ik}$ as the normalized Euclidean distances between centroids $c_i$ and  $c_k$ of cells $i$ and $k$ respectively, as:
\begin{equation}
C_{ik} = \frac{\|c_i - c_k\|_2}{d_{\max}}, \quad d_{\max} = \max_{p,q} \|c_p - c_q\|_2
\label{eq:ground-cost}
\end{equation}
This normalization ensures $C_{ik} \in [0, 1]$, rendering the $W_1$ invariant to the absolute spatial scale. This allows an objective comparison across different geographic regions or different grid resolutions without bias from physical units (meters vs. kilometers).
We solve this optimization problem using the exact solver provided by the Python Optimal Transport library~\cite{flamary2021pot}.
\paragraph{Kendall rank correlation coefficient~\cite{kendall1938new}}
\label{appendix:def_kendalltau}
Let $X$ and $Y$ be two rankings (permutations) over subsets of the same universe $U$ of items. To compare rankings with non-identical supports, we extend both rankings to the union of their supports, denoted as $\mathcal{S} = \text{supp}(X) \cup \text{supp}(Y)$. Let $n = |\mathcal{S}|$ be the total number of unique items.
Items missing from a list are assigned a tied rank at the bottom of that list.
The Kendall rank correlation coefficient ($\tau_b$), adjusted for ties, is defined as:
\begin{equation}
\label{equation:kendall}
\tau_b(X,Y) = \frac{N_c - N_d}{\sqrt{(N0-T_x)(N_0-T_y)}}
\end{equation}
where:
\begin{itemize}
    \item $N_c$ is the number of \textit{concordant pairs}. A pair $(i, j)$ is concordant if the relative ordering of $i$ and $j$ is the same in both $X$ and $Y$.
    \item $N_d$ is the number of \textit{discordant pairs}. A pair is discordant if the ordering is inverted.
    \item $N_0 = \binom{n}{2} = \frac{n(n-1)}{2}$ is the total number of item pairs.
    \item $T_x$ is the number of tied pairs in $X$ only.
    \item $T_y$ is the number of tied pairs in $Y$ only.
\end{itemize}
The Kendall rank coefficient is often used as a test statistic in a statistical hypothesis test to establish whether two variables may be regarded as statistically dependent. The Kendall's $\tau_b$ ranges from $[-1,1]$, where $1$ indicates that $X$ and $Y$ have identical ranking (perfect correlation), $0$ indicates no correlation, and $-1$ indicates a complete inversion of the rankings.
Unlike the standard $\tau_a$ (which does not account for ties), $\tau_b$ is robust to the large number of ties introduced by the missing item handling strategy.

\subsubsection{I-rank}
The Individual rank (I-rank)~\cite{gonzalez2008understanding,pappalardo2013understanding}, characterizes the visitation frequency ordering of locations for each individual trajectory/user.
To evaluate the preservation of these hierarchies, we compare the distributions of I-rank vectors of the synthetic dataset to those of the original dataset.

After discretizing all trajectories (see appendix~\ref{appendix:Discretization} for discretization), we project each user's rank vector into a scalar using the Kendall rank correlation coefficient, $\tau_b$, as defined in equation~\ref{equation:kendall}.

Specifically, for a user $u$, with a trajectory $\mathcal{T}_u$, we compute a normalized distance $d_u \in [0,1]$, defined as:
\begin{equation}
    d_u = \frac{1-\tau_b(I_u, I_0)}{2}
\end{equation}
where $I_u$ represents the vector of visitation frequencies for locations in $T_u$, and $I_0$ is a reference vector containing the corresponding spatial location (grid indexes), sorted by index.
This normalization maps perfect spatial concordance to $0$, independence to $0.5$, and complete inversion to $1$.
We note that all datasets are discretized using the same spatial grid and cell indexing, so that the reference permutation $I_0$ is shared across real and synthetic users.

Finally, the fidelity of the synthetic dataset's I-ranks is quantified by the 1-Wasserstein distance $W_1$, as defined in~\ref{appendix:def_wasserstein} between the empirical distributions of these projected scores.

\subsubsection{Pairwise distances}
To evaluate the relational statistics preservation at the trajectory level, we analyze the distribution of pairwise similarity distances~\cite{biagioni2012inferring} within each dataset.

We employ two distinct distance metrics to capture the multi-dimensional nature of mobility traces:
\begin{itemize}
    \item The Hausdorff distance~\cite{huttenlocher2002comparing}, which measures the spatial proximity between two trajectories by calculating the maximum distance from a point in one trace to the nearest point in the other. Comparing the distributions of pairwise Hausdorff distances allows the assessment of how well the synthetic data preserves the spatial density and geometric spread of the original trajectories.
    \item The cosine distance applied to the PoI category dimension to evaluate the semantic similarity between the trajectory within a dataset.
\end{itemize}

For each synthetic dataset, $D_{\text{syn}}$, we compute these distances for all possible pairs of trajectories, resulting in a distribution of distances.

Each distribution is then compared with the distribution computed on the associated original dataset, $D$ via the 1-Wasserstein distance, $W_1$.
\subsubsection{G-rank}
The G-rank (Global Rank) ~\cite{ouyang2018non} represents the frequency-based ordering of locations, at the point-level. We apply the G-rank on discretized trajectories and quantify the similarity between the original ($D$) and synthetic ($D_{syn}$) dataset via Kendall's $\tau_b$.
\subsubsection{Semantic importance: Categorical G-rank}
We evaluate the preservation of the semantic importance, at the point-level by utilizing the G-rank ~\cite{ouyang2018non} on the categories, instead of geographic locations, representing the frequency-based ordering of PoI categories across the entire dataset. This statistics captures the popularity distribution of different activity types (e.g., Restaurant vs. Shops).

We quantify the similarity between the Categorical G-ranks of the original ($D$) and synthetic ($D_{\text{syn}}$) datasets, using Kendall's $\tau_b$ to penalize the synthetic data if it misrepresents the relative importance of specific activity hubs (e.g., incorrectly ranking leisure activities as more frequent than residential stays).
\subsubsection{Average speed}
\label{appendix:metric-averagespeed}
For each trajectory $T =[(p_1,t_1),\cdots,(p_{|T|}, t_{|T|})]$ with positions expressed in a projected metric CRS, we define step speeds as $v_i = \|p_{i+1} - p_i\|_2 / (t_{i+1} - t_i)$ for $i$ such that $(t_{i+1} - t_i) > 0$.

We consider the per‑trajectory average speed distribution, resulting in a one-dimensional distribution of per trajectory average speed for a dataset.

We compare the distributions $V$ and $V_{\text{syn}}$ for the original and synthetic datasets, $D$ and $D_{\text{syn}}$ respectively. We quantify discrepancy with the 1‑Wasserstein distance $W_1(V, V_{\text{syn}})$, as defined in equation~\ref{eq:1w} using the absolute difference ground cost over km/h.
\subsubsection{Transition probability}
\label{appendix:metric-transitionprob}
We evaluate the preservation of point-level relational mobility statistics by comparing first-order spatial transition kernels~\cite{meyn2012markov} between the original and synthetic datasets. For each dataset, we construct a row stochastic first-order transition matrix by aggregating observed transitions between discretized spatial states across all trajectories (see ~\ref{appendix:Discretization} for grid discretization). Each entry $p_{ij}$ represents the empirical conditional probability of transitioning to cell $j$ given current cell:
\begin{equation}
p_{ij} = \frac{\text{count}(s_i \to s_j)}{\sum_{k \in S} \text{count}(s_i \to s_k)}, \quad \forall s_i, s_j \in S
\end{equation}
where $S$ is the discrete state space of the grid.
This yields a population-level Mobility Markov Chain (MMC)~\cite{gambs2010show}, where each row $P_i$ represents the conditional distribution of next states given an origin $i$.

We compare the resulting conditional distributions from the original and synthetic datasets with $W_1$~\ref{eq:1w}, employing the spatial ground cost matrix defined in Eq.~\ref{eq:ground-cost} to accounts for physical displacement between cells.
\begin{equation}
d_i =
\begin{cases}
W_1(P_i, P_i^{\text{syn}}) & \text{if } \sum P_{i}^{\text{syn}} > 0 \\
1 & \text{otherwise}
\end{cases}
\end{equation}

If the synthetic data has no outgoing mass from an origin state present in the real data, we assign the maximal penalty $d_i = 1$. The final discrepancy measure aggregates per-state distances using visitation frequencies $\pi_i$, from the original dataset as weights:

\begin{equation}
    D(P, P^{\text{syn}}) = \sum_{i \in S} \pi_id_i, \quad \pi_i = \frac{n_i}{\sum_{k \in S} n_k},
\end{equation}
where $n_i$ is the number of observed transitions originating in state $i$.
This means that if the synthetic data produces transitions from a state $i$ that never existed in the real data, its weight $w_i$ will be 0, and it will be effectively ignored.
Weighting by the true state distribution is standard when defining expected risk for probabilistic models under the true data distribution. This provides a fixed evaluation across generators and reduces variance from sparsely observed states~\cite{gneiting2007strictly}. Because $C \in [0,1]$ and $\sum_i \pi_i = 1$, the error $D$ is unitless and lies in $[0,1]$.
\subsection{Realism assurance}
\subsubsection{Trajectory spatial implausibility}
This metric evaluates the geographic realism of synthetic trajectories by identifying points situated in \textit{impossible} locations. We define a spatial implausibility as a state where a trajectory point resides within a body of water or a restricted area (e.g., a military base).
Because, GPS has intrinsic noise, we apply a second constraint: we check if the implausible point maintains a significant distance from any known traversable infrastructure (roads, bridges, piers, or buildings). This enables the differentiation between genuine implausibility and artifacts caused by GPS noise.

We define the set of accessible land infrastructure $\mathcal{L}$ as the union of building footprints and road infrastructure. The Implausible domain is denoted by $\mathcal{I}$. All spatial features, used to label points, are extracted from the OpenStreetMap (OSM) service~\cite{vargas2020openstreetmap}.

Formally, a point $p$ violates a spatial constraint if:
\begin{equation}
    \mathbb{V}(p) = \begin{cases} 1, & \text{if } p \in \mathcal{I} \text{ and } \text{dist}(p, \mathcal{L}) > \delta \\ 0, & \text{otherwise} \end{cases}
    \label{eq:implausibility}
\end{equation}

where $\delta$ represent a tolerance radius, calibrated to account for the heteroskedasticity of GPS error. We set $\delta =30 \text{ meter}$, which is the standard GPS spatial imprecision in urban areas.
We recognize a Non-Zero Baseline Error as empirical real trajectory data often exhibits a non-zero violation rate ($\epsilon$) due to signal multipath effects in urban canyons.
We define the Implausibility Rate as the proportion of trajectories in a dataset that exhibit at least one constraint violations.
\subsubsection{Location implausibility}
\label{appendix:location_implaus}
To supplement the trajectory-level analysis, we evaluate geographic plausibility at the point level. This point-wise metric leverages the same logic defined in equation~\ref{eq:implausibility} to report the global ratio of implausible locations across all generated points.
\subsubsection{Category-location match}
\label{appendix:semantic_implaus}
This metric is intended to assess how well the synthetic data preserves the dominant semantic label (e.g., restaurant, shop) associated with each spatial cell grid in the real data.
Intuitively, if a cell is overwhelmingly tagged as restaurant cell in the real data, its synthetic counterpart should also be dominated by restaurant. To ensure statistical robustness, we exclude cells where the data is too sparse to establish a pattern or where no single category clearly dominates. We define these as eligible cells, $S^*$, based on a minimum $k_{\text{min}}$ observation threshold and a dominance ratio, $\delta$ (the share of the top category relative to the total). In our evaluation we set $k_{\text{min}}=5$ and $\delta=0.5$

We discretize the study area into a uniform grid as detailed in Appendix~\ref{appendix:Discretization}. For a dataset $X \in \{D, D_{\text{syn}}\}$, the dominant (top-1) category $k^*(s^*)$ for a cell $s^* \in S^*$ is:

\begin{equation}
k^*(s^*) = \arg\max_{k \in K} n_X(s^*, k)
\end{equation}
where $n_X(s^*,k)$ is the count of observations for category $k$ in cell $s^*$. The match rate is calculated as the accuracy of the top-1 category across the set of eligible cells $S^*$:
\begin{equation}
    \mathrm{MR} = \frac{1}{|S^*|} \sum_{s^* \in S^*}, \mathbbm{1}_{K}
\quad \text{with}
    \mathbbm{1}_{K} =
    \begin{cases}
    1 & \text{if } k^{*}(s^*) = k^{*}_\text{ syn}(s^*),\\ 0, & \text{otherwise}
    \end{cases}
\end{equation}
\subsubsection{Map reconstruction}
This metric evaluates the deviation of synthetic trajectories from the physical road infrastructure.
Since most generative models operate in a continuous coordinate space, verifying adherence to the physical infrastructure is therefore crucial to ensuring realism.

We define $G = (V, E)$ the graph of the partition of the urban road network activated by the original dataset. $E$ is the set of the graph edges representing the roads segments. $E$ is constructed by mapping each point $p \in D$ to its nearest edge, $e$ in the global network infrastructure of the study area. $V$ is the set of nodes delimiting the end of the start of every edge. We worked with a network infrastructure from the Openstreetmap (OSM)~\cite{vargas2020openstreetmap} map service.

Given the inherent sparsity of mobility datasets, $E = \{e_i\}$ is a collection of topologically disconnected segments. To reconstruct a latent continuous infrastructure, we perform a shortest-path imputation between consecutive observed edges: For a trajectory transition $(e_t,e_{t+1})$, we compute the shortest path, $P_t$ for a point at the segment $e_t$ to reach the road segment $e_{t+1}$ using the Dijkstra's algorithm ~\cite{lou2009map}.
If no path was found, (representing a topological jump), the transition preserved as is.

The Empirical Infrastructure $\mathcal{I}$ is defined as the union of all observed edges and imputed paths, $\mathcal{I} = E \cup \bigcup_{t} P_t$.

For each point in the synthetic dataset, $D_{\text{syn}}$, we calculate the Euclidean distance to the nearest segment in $\mathcal{I}$. We report the average displacement in kilometers, aggregated first at the trajectory level and then across the entire synthetic dataset $D_{\text{syn}}$.
Note that for computational reasons, we restrict the analysis to highways and filter the movements on highway roads (points labeled as high-speed movements)
\subsection{Task performance}
Beyond verifying statistics preservation and the realism of the synthetic data, we evaluate if it can effectively substitute for real data for a downstream task.
To achieve this, we employ a Train on Synthetic, Test on Real (TSTR) protocol~\cite{esteban2017real, jordon2018pate}. It evaluates whether the mobility dynamics learned by the generative model are sufficiently robust to predict unseen human movement in the physical study area.
We consider three predictive mobility tasks that we present below, both at the trajectory and point level:
\subsubsection{Trajectory next location prediction}
To evaluate individual-level predictability, we implement the Mobility Markov Chain (MMC) framework proposed by~\citet{gambs2012next}.
Given the absence of a one-to-one mapping between synthetic and real trajectories, we derive a collective synthetic transition operator $P^{\text{syn}} \in \mathbb{R}^{n \times n}$ defined in~\ref{appendix:metric-transitionprob} as a predictive model. Each entry $P_{ij}$ represent the maximum likelihood estimate of the probability of transitioning from cell $i$ to cell $j$.
For rows where the synthetic generator produces no outgoing mass (sink states), we assume a uniform distribution over the state space to ensure a well-defined predictor.

We evaluate the predictive power of $P_{\text{syn}}$ on each trajectory in the original dataset $D$, we perform a step-wise prediction: given a current state $s_n$, the model predicts a candidate set of future states $\hat{S}_{n+1}$ consisting of the $k$ most probable subsequent states for $n+1$ step from $P_i^{\text{syn}}$.

We evaluate this predictor on each real trajectory $T = (s_1, s_2, \dots, s_L)$ in the original dataset $D$. The accuracy for a single trajectory of length $L$ is defined as:
\begin{equation}
Acc(T) = \frac{1}{L-1} \sum_{n=1}^{L-1} \mathbb{I}(s_{n+1} \in \hat{S}{n+1})
\end{equation}
where $\mathbb{I}(\cdot)$ is the indicator function. We report the mean top-k accuracy across all trajectories,$\frac{1}{|D|} \sum_{T \in D} \text{Acc}(T)$. In our evaluation we report $k=10$.

\begin{table*}[htbp!]
\centering
\begin{subtable}[t]{\textwidth}
\centering
\begin{tabular}{
    l
    l
    c
    c
    c
    c
    c
}
\toprule
    &
    & I-rank
    & Transition probabilities
    & Global Flow Prediction\\
    & & ($W_1$)  & ($W_1$)  & ($W_1$) \\
    & \textbf{Original}  & 0.000 $\pm$ 0.000   & 0.000 $\pm$ 0.000  & 0.015 $\pm$ 0.001 \\
    \cmidrule(lr){2-5}
    & \textbf{LSTM-TrajGAN~\cite{rao2020lstm}}   & 0.101 $\pm$  0.030 &  0.230 $\pm$ 0.140 & \textbf{0.027 $\pm$ 0.002}\\
    & \textbf{exGAN~\cite{song2023except}}     & 0.132 $\pm$  0.025  &   \textbf{ 0.214 $\pm$0.150} &  0.030 $\pm$ 0.003\\
    & \textbf{TrajGDM~\cite{chu2024simulating}}    & \textbf{0.023 $\pm$ 0.010}  & 0.592$\pm$0.236 & 0.051$\pm$ 0.007\\
\bottomrule
\end{tabular}
\caption{Stability analysis on Weekly FS-NYC~\cite{may2020marc}.}
\end{subtable}
\hfill
\begin{subtable}[t]{\textwidth}
\centering
\begin{tabular}{
    l
    l
    c
    c
    c
    c
    c
}
\toprule
    &
    & G-rank
    & Transition probabilities
    & Global flow prediction\\
    & &  ($\tau_b$)  & ($W_1$) & ($W_1$) \\
    & \textbf{Original}  & 1.000 $\pm$  0.000   & 0.000 $\pm$  0.000  & 0.008 $\pm$  0.000 \\
    \cmidrule(lr){2-5}
    & \textbf{LSTM-TrajGAN~\cite{rao2020lstm}}  & \textbf{0.051 $\pm$ 0.186}  & \textbf{0.154 $\pm$ 0.118}   & \textbf{0.012 $\pm$ 0.001} \\
    & \textbf{exGAN~\cite{song2023except}}    & 0.042$\pm$ 0.183 & 0.164 $\pm$ 0.122 & \textbf{0.012 $\pm$  0.001} \\
    & \textbf{TrajGDM~\cite{chu2024simulating}}  & -0.175 $\pm$ 0.189  & 0.469 $\pm$ 0.279  & 0.025 $\pm$ 0.005\\
\bottomrule
\end{tabular}
\caption{Stability analysis on Daily Geolife}
\end{subtable}
\caption{Stability analysis on the grid-based metrics used for the utility evaluation (on the Weekly FS-NYC and the daily Geolife datasets)~\cite{may2020marc}.}

\label{tab:stabilityanalysis}
\end{table*}

\subsubsection{Traffic flow prediction}
We further define a traffic flow prediction task as a a Global Mobility Markov Chain (GMMC)~\cite{gambs2010show}, to evaluate point-level task performance. In this task, we treat the synthetic transition matrix $P^{\text{syn}}$, constructed in~\ref{appendix:metric-transitionprob} as a linear operator that models population-level flux.
Rather than predicting individual paths~\cite{gambs2012next}, we evaluate the evolution of the population spatial distribution over the discrete state space.
Following the TSTR protocol, we apply the transition matrix $P^{\text{syn}}$ to the empirical population distributions observed in the real dataset $D$: we represent the normalized distribution of all users at sequence step $n$ in the original data by $V_n$.
For rows where the synthetic generator produces no outgoing mass (sink states), we assume a uniform distribution over the state space.
The predicted distribution at the subsequent step is calculated as $\hat{V}_{n+1}=V_nP^{\text{syn}}$ and compared against the ground truth $V_{n+1}$ using the $W_1(\hat{V}_{n+1},V_{n+1})$ with a spatial ground cost~\ref{appendix:def_wasserstein}. We average this discrepancy across the first $N$ sequence steps, where $N$ is the 90th percentile of trajectory lengths to ensure statistical stability.
\subsubsection{Trajectory clustering}
To evaluate the predictive utility of the synthetic trajectories at the trajectory-level, we perform a density-based clustering task. We first define a "spatial world view" by fitting a density-based clustering (HDBSCAN)~\cite{campello2013density} on the geographic centroids of all trajectories in the synthetic data $D_{\textbf{syn}}$. This process extracts the latent spatial density manifold and identifies global centers of activities (geographic regions where individual movements tend to gravitate) within the synthetic datasets.

We then evaluate the generalization of these synthetic structures on the original data: unseen real-world trajectory centroids are projected onto this manifold and assigned to the nearest synthetic cluster using the approximate membership protocol \cite{McInnes2017hdbscan}.
We quantify the utility of the learned synthetic boundaries using the \textit{Silhouette score}.
For a trajectory centroid, $c_T$ in the original dataset $D$, the silhouette is defined, $s(c_T) \in [-1, 1]$ is defined as:
\begin{equation}
s(c_T) = \frac{b(c_T) - a(c_T)}{\max\{a(c_T), b(c_T)\}}
\label{eq:silhouetescore}
\end{equation}
where $a(c_T)$ is the mean distance between trajectory centroid $c_T$ and all other trajectory centroids in the same assigned synthetic cluster and $b(c_T)$ is the distance to nearest neighboring cluster.
A high Silhouette score indicates that the synthetic model has successfully captured the distinct spatial separation of actual activity zones.
We report the mean Silhouette score across all trajectories.
\section{Geospatial Discretization}
\label{appendix:Discretization}
To facilitate the computation of certain metrics on trajectory data, we discretize the continuous geographic space. The representation of trajectories as sequences of latitude-longitude coordinates can make metrics like spatial density or transition probability challenging to compute directly. To address this, we partition the geographic space into a uniform, absolute grid~\cite{teixeira2019deciphering, lin2012predictability} anchored at a fixed global origin, $(0,0)$ in the equivalent metric Coordinate Reference System (CRS), where each record was assigned to the cell of the grid within which it lies and was represented by a unique cell identifier.
\subsection{Grid size selection}
For mobility data, grid cell sizes on the order of 500 $m^2$ to 2 $km^2$ is considered consistent with resolutions used in urban hotspot analysis~\cite{louail2014mobile} while 100 $m^2$ cell size for traffic analysis~\cite{liu2017grid}. Since in our evaluation framework requires a robust state space for multiple downstream metrics, we opt for a single grid cell size per dataset that balances the trade-off between precision and computation efficiency.

To select a cell grid size for each of our four datasets, we analyze the impact of the spatial resolution on three quantitative properties of the discretized trajectories~\cite{smolak2021impact}:
\begin{inparaenum}[a)]
\item unique transitions (the fraction of unique origin-destination pairs),
\item self-transitions (the fraction of movements contained within a single cell), and
\item spatial occupancy (the ratio of visited cells to the total grid space)
\end{inparaenum}
We define a search space based on the segment length (i.e., distance between two consecutive points) where the lower and upper bounds are determined by the 10th and 50th percentiles respectively of the segment distribution. This ensures the resolution remains fine enough to capture the smallest meaningful movements while filtering out sub-resolution sensor noise.

By identifying the inflection points (elbows) of the three properties independently, we can detect the regime where marginal gains in resolution no longer justify the exponential increase in sparsity, thereby narrowing the candidates to a feasible region of Pareto frontier.
The final grid size is selected as the consensus midpoint of these independent candidates. This compromise ensures the discretization retains sufficient resolution while providing a robust, non-sparse state space for Markovian mobility modeling. For our datasets, this yields a cell edge lengths of 500m for the weekly FS-NYC and 350m for daily Geolife respectively. For the additional datasets weekly and hourly Geolife, the selected edge cell lengths are 600m and 100m, respectively.

These are the values used in the runs that produced Tables~\ref{tab:usecaseA}-~\ref{tab:usecaseB} (and tables~\ref{tab:usecaseBWG}–~\ref{tab:usecaseBHG})
\subsection{Spatial stability analysis}
The choice of the cell size (spatial resolution) affects the granularity of the discretized trajectories and therefore the resulting statistics. This is a classic challenge in spatial analysis, formally known as the Modifiable Areal Unit Problem (MAUP)~\cite{fotheringham1991modifiable}, which states that analytical outcomes are dependent on the scale (size) and zoning (shape) of the spatial units used. For this reason, an optimal cell size for one statistic (e.g., spatial density) may not be optimal for another (e.g., transition probabilities).
Given this sensitivity, we performed a stability analysis. We computed the grid-based statistics across multiple grid configurations, varying the cell size from 100 meter to 1000 meters in 50-meter increments. For each cell size, we generated a set of grid phase shifts (offset from the grid origin) that span the full equivalence class of lattice alignments. Table~\ref{tab:stabilityanalysis} summarizes these findings, reporting the mean and standard deviation for each grid-based statistics we used in our utility evaluation to verify the robustness of our conclusions.

\section{Evaluation Datasets statistical properties}
\label{appendix:properties}
To enhance diversity, we derive four evaluation sets: Weekly FS‑NYC and three Geolife variants created by temporal aggregation and down-sampling to control sequence length.
\paragraph{Temporal variants} We split the Geolife dataset into three variants based on three temporal aggregations: week of the month, day of the week and hour of the day. We down-sampled trajectories, to limit the sequences lengths. This yields different sampling densities, ranging from fine-grained (hourly) to coarse grained (weekly) and correspondingly  different trajectory statistics as shown in table~\ref{tab:dataset_stats}, effectively transforming a single large dataset into 3 multiple statistically diverse datasets, each presenting a unique learning challenge.
Table~\ref{tab:dataset_stats} presents the following statistical descriptors:
\begin{inparaenum}
    \item $\mu_{\Delta t}=\mathbb{E}[t_{i+1}-t_i]$ is the mean sampling interval in minutes, 
    \item $CV_{\Delta t}=\mathrm{std}(\Delta t)/\mu_{\Delta t}$ is the relative variation of the sampling interval, which measures temporal irregularity ($CV=$0 represents perfect regularity whereas $CV>1$ represents gaps),
    \item $P(\Delta t>\tau_{\text{gap}})$ is the proportion of temporal extreme gaps (gaps exceeding a fixed threshold $\tau_{\text{gap}}$: set to $10\times$ the median $\Delta t$),
    \item \#Traj denotes the number of trajectories,
    \item $\mu_{\text{Length}}$ and $95\%_{\text{Traveled distance}}$ are respectively the median and 95th percentile of the trajectory lengths (number of points),
    \item $\mu_{\text{Traveled distance}}$ is the mean traveled distance and $\mu_{\text{displacement}}$ is the mean displacement e.g., distance between two consecutive points (in kilometers).
\end{inparaenum}
\paragraph{Semantic annotation} We enrich locations with coarse semantic categories to support semantic utility metrics and the semantic input dimension of the LSTM-TrajGAN and exGAN models. We leveraged Openstreetmap (OSM) tags~\cite{vargas2020openstreetmap} to assign a category to stay points (when individuals are stationary) and slow-movement points. We map OSM features to 9 categories: administrative, commercial, education, healthcare, leisure, natural, residential, transportation infrastructure, other. The tenth category represents high‑speed segment, labeled as on‑road/vehicle
\begin{table*}[htbp!]
\centering
\begin{tabular}{lccccccccc}
\toprule
\textbf{Dataset} 
& $\boldsymbol{\mu_{\Delta t}}$ 
& $\boldsymbol{CV_{\Delta t}}$ 
& $\boldsymbol{P(\Delta t>\tau_{\text{gap}})}$ 
& \textbf{\#Traj} 
& \textbf{$\mu_{\text{Length}}$}
& \textbf{$95\%_{\text{Traveled distance}}$}
& \textbf{$\mu_{\text{Traveled distance}}$}
& \textbf{$\mu_{\text{displacement}}$} \\
\midrule
Weekly FS-NYC   & 360 & 0.5 & 0.20 & 3079 & 17 & 51 & 72.8 & 5.8  \\
Weekly Geolife  & 300 & 0.5 & 0.12 & 1752 & 20 & 57 & 81.4 & 5 \\
Daily Geolife   & 38 & 0.3 & 0.07 & 7706 & 13 & 49 & 26.8 & 4.8 \\
Hourly Geolife  & 0.6 & 0.4 & 0.01 & 44287 & 43 & 122 & 6.2 & 4  \\
\bottomrule
\end{tabular}
\caption{Statistical descriptors of the different trajectory datasets used in this work.}
\label{tab:dataset_stats}
\end{table*}

\section{Additional Datasets}
\label{appendix:add_datasets}
 Table~\ref{tab:usecaseBWG} and ~\ref{tab:usecaseBHG} summarizes the results of our utility evaluation on the weekly and hourly splits of the Geolife dataset.

 \begin{table*}[!htbp]
\centering
\begin{tabular}{
    l
    l
    c
    c
    c
    c
}
\toprule
    & \textit{Weekly Geolife}
    & \multicolumn{2}{c}{\textbf{Statistics preservation}}
    & \multirow{2}{*}{\textbf{Realism assurance} }
    & \multirow{2}{*}{\textbf{Task performance}} \\
    
    & \textit{(Use-case B)}
    & \textit{Marginal statistics}
    & \textit{Relational statistics}
    & & \\
    \midrule
    \multirow{6.5}{*}{\rotatebox{90}{Trajectory level}}
    &
    & Average speed
    & Pairwise Hausdorff
    & Map reconstruction
    & Next location prediction \\
    & & ($W_1$ (km/h)) & ($W_1$) & (mean (km)) & (Mean acc) \\
    & \textbf{Original}  & 0.000   & 0.000 & 0.055 & 0.528  \\
    \cmidrule(lr){2-6}
    & \textbf{LSTM-TrajGAN~\cite{rao2020lstm}}   & 0.242  &  \textbf{0.001} & \textbf{0.147} & \textbf{0.273}\\
    & \textbf{exGAN~\cite{song2023except}}     & \textbf{0.113} & 0.002  & 0.187 & 0.115\\
    & \textbf{TrajGDM~\cite{chu2024simulating}}    & N/A   & 0.074  & 0.272 & 0.155 \\
\midrule
    \multirow{6.5}{*}{\rotatebox{90}{Point level}}
    &
    & G-rank
    & Transition probabilities
    & Location implausibility
    & Global flow prediction \\

    &  & ($\tau_b$) & ($W_1$) & (Ratio) & ($W_1$) \\
    & \textbf{Original}  & 1.000 & 0.000 & 0.002 & 0.005 \\
    \cmidrule(lr){2-6}
    & \textbf{LSTM-TrajGAN~\cite{rao2020lstm}}  & \textbf{0.206} & \textbf{0.090} & \textbf{0.016}  & \textbf{0.010} \\
    & \textbf{exGAN~\cite{song2023except}}     & 0.039 
    & 0.186  & 0.024 & 0.018\\
    & \textbf{TrajGDM~\cite{chu2024simulating}}   & -0.025 & 0.300 &  0.018 & 0.019 \\
\bottomrule
\end{tabular}
\caption{Results of the application of our framework on the generated data (Weekly Geolife).}

\label{tab:usecaseBWG}
\end{table*}

 \begin{table*}[!htbp]
\centering
\begin{tabular}{
    l
    l
    c
    c
    c
    c
}
\toprule
    & \textit{Hourly Geolife}
    & \multicolumn{2}{c}{\textbf{Statistics preservation}}
    & \multirow{2}{*}{\textbf{Realism assurance} }
    & \multirow{2}{*}{\textbf{Task performance}} \\
    
    & \textit{(Use-case B)}
    & \textit{Marginal statistics}
    & \textit{Relational statistics}
    & & \\
    \midrule
    \multirow{6.5}{*}{\rotatebox{90}{Trajectory level}}
    &
    & Traveled distance
    & Pairwise Hausdorff
    & Map reconstruction
    & Next location prediction \\
    & & ($W_1$ (km)) & ($W_1$) & (mean (km)) & (Mean acc) \\
    & \textbf{Original}  & 0.000  & 0.000 & 0.045 & 0.845  \\
    \cmidrule(lr){2-6}
    & \textbf{LSTM-TrajGAN~\cite{rao2020lstm}}   & 4.864  &  0.002 &  0.124  & 0.178 \\
    & \textbf{exGAN~\cite{song2023except}}     & \textbf{3.881} & \textbf{0.001} & \textbf{0.118} & \textbf{0.279}\\
    & \textbf{TrajGDM~\cite{chu2024simulating}}  & 234.176  & 0.066  & 0.852 & 0.007\\
\midrule
    \multirow{6.5}{*}{\rotatebox{90}{Point level}}
    &
    & G-rank
    & Transition probabilities
    & Location implausibility
    & Global flow prediction \\

    &  & ($\tau_b$) & ($W_1$) & (Ratio) & ($W_1$) \\
    & \textbf{Original}  & 1.000 & 0.000 & 0.004 & 0.001 \\
    \cmidrule(lr){2-6}
    & \textbf{LSTM-TrajGAN~\cite{rao2020lstm}}  & -0.217 & 0.226   & \textbf{0.020} & \textbf{0.020}\\
    & \textbf{exGAN~\cite{song2023except}}     &  -0.169 & \textbf{0.182} & \textbf{0.020} & 0.021 \\
    & \textbf{TrajGDM~\cite{chu2024simulating}}   & -0.369  & 0.970 &  0.055 & 0.053 \\
\bottomrule
\end{tabular}
\caption{Results of the application of our framework on the generated data (Hourly Geolife).}

\label{tab:usecaseBHG}
\end{table*}